\documentclass[number,preprint,3p]{elsarticle}

\usepackage{multirow}
\usepackage{color}
\usepackage{graphicx}
\usepackage{subcaption}
\usepackage{algorithm}
\usepackage{bm}
\usepackage[colorlinks]{hyperref}
\usepackage{amssymb}
\usepackage{amsthm}
\usepackage{amsmath}
\usepackage{booktabs}
\usepackage{url}
\usepackage{scrextend}
\usepackage{epstopdf}
\usepackage{float}
\usepackage{tablefootnote}
\usepackage{mathtools}
\usepackage{soul}
\usepackage[table]{xcolor}
\usepackage[perpage]{footmisc}
\usepackage{lineno}

\makeatletter
\gdef\urlauthor#1#2{\g@addto@macro\@elsuads{\let\corref\@gobble%
     \def\@@tmp{#1}\raggedright\eadsep
     {\ttfamily\url{\expandafter\strip@prefix\meaning\@@tmp}}\space(#2)%
     \def\eadsep{\unskip,\space}}%
}
\gdef\emailauthor#1#2{\stepcounter{ead}%
     \g@addto@macro\@elseads{\raggedright%
      \let\corref\@gobble\def\@@tmp{#1}%
      \eadsep{\ttfamily\href{mailto:\expandafter\strip@prefix\meaning\@@tmp}{\expandafter\strip@prefix\meaning\@@tmp}}
      (#2)\def\eadsep{\unskip,\space}}%
}
\makeatother

\def\r{\mathbb{R}}

\makeatletter
\let\@afterindenttrue\@afterindentfalse
\makeatother
\makeatletter
\patchcmd{\ps@pprintTitle}
  {Preprint submitted to}
  {Preprint submitted to}
  {}{}
\makeatother

\biboptions{numbers,sort&compress,square}
\journal{arXiv}

\begin{document}
\begin{frontmatter}
\renewcommand{\thefootnote}{\fnsymbol{footnote}}

\title{Constructing Balanced Datasets for Predicting Failure Modes in Structural Systems Under Seismic Hazards}

 \author[1]{Jungho Kim}
 \author[2]{Taeyong Kim\corref{cor1}}
\ead{taeyongkim@ajou.ac.kr}  
         \cortext[cor1]{Corresponding author}
 \address[1]{Department of Civil and Environmental Engineering, University of California, Berkeley, United States}
 \address[2]{Department of Civil Systems Engineering, Ajou University, Suwon, Republic of Korea} 
\begin{abstract}
Accurate prediction of structural failure modes under seismic excitations is essential for seismic risk and resilience assessment. Traditional simulation-based approaches often result in imbalanced datasets dominated by non-failure or frequently observed failure scenarios, limiting the effectiveness in machine learning-based prediction. To address this challenge, this study proposes a framework for constructing balanced datasets that include distinct failure modes. The framework consists of three key steps. First, critical ground motion features (GMFs) are identified to effectively represent ground motion time histories. Second, an adaptive algorithm is employed to estimate the probability densities of various failure domains in the space of critical GMFs and structural parameters. Third, samples generated from these probability densities are transformed into ground motion time histories by using a scaling factor optimization process. A balanced dataset is constructed by performing nonlinear response history analyses on structural systems with parameters matching the generated samples, subjected to corresponding transformed ground motion time histories. Deep neural network models are trained on balanced and imbalanced datasets to highlight the importance of dataset balancing. To further evaluate the framework’s applicability, numerical investigations are conducted using two different structural models subjected to recorded and synthetic ground motions. The results demonstrate the framework's robustness and effectiveness in addressing dataset imbalance and improving machine learning performance in seismic failure mode prediction.
\end{abstract}
\end{frontmatter}
\renewcommand{\thefootnote}{\fnsymbol{footnote}}

\section{Introduction}

\noindent The widespread adoption of data-driven methodologies marks a significant paradigm shift from traditional theory- and mechanics-based approaches to exploratory- and evidence-based strategies. This new paradigm leverages large-scale datasets, advanced computational tools, and machine learning techniques to uncover hidden patterns and generate predictive insights that were previously unattainable \cite{lu2022ai,kim2024deep}. In the field of earthquake engineering, this shift has enabled proactive risk management by accelerating predictions of structural responses and damage states under seismic excitations \cite{oh2024deep, zhang2024transformer,kim2025efficient}, identifying vulnerabilities to prioritize mitigation strategies \cite{mojtahedi2017critical,kalakonas2022seismic,wang2023assessment}, and providing quantitative insights into seismic performance and risk \cite{ellingwood2009quantifying,cremen2021decision,amini2024automated,kim2025uncertainty}. Furthermore, it supports the optimal design of structural systems to ensure resilience against seismic hazards \cite{hao2023towards,kim2024active,tan2025intelligent}.

The prompt prediction of failure modes in structural systems, in addition to their seismic responses, represents another critical application of data-driven methodologies \cite{mangalathu2019machine,naderpour2021failure,kabir2021failure,kim2025near}. This capability is crucial for determining optimal recovery strategies immediately following a seismic event. Despite their potential, such machine learning methods face a significant challenge, particularly in predicting ``black swan" failure modes due to imbalanced data categories during model training. The ``black swan" refers to an event that is highly improbable yet carries catastrophic consequences. When a dataset contains disproportionately fewer samples for rare failure modes compared to more common ones, machine learning models trained on such data may become biased towards more frequent categories. This imbalance reduces the model’s predictive performance for less common but critical failure modes, posing a significant limitation to achieving the seismic resilience of structural systems.

Various methods have been developed to address the issue of imbalanced datasets \cite{kim2022machine,jafarigol2023review,zhou2023machine,yuan2024learning,qu2025two}, which can be broadly categorized into three. First, resampling techniques such as under- or over-sampling can be used to balance the class ratios by adjusting the representation of the minority or majority classes. Second, algorithmic modifications, wherein classification algorithms are adjusted to minimize the misclassification error for the minority class, can be employed. Note that these two approaches primarily address imbalance issues in pre-constructed datasets. Third, during dataset construction, a proactive strategy can be employed by identifying multiple failure modes and ensuring proper sampling of each mode to create a balanced dataset. Advanced techniques for this purpose include genetic algorithms \cite{kim2013system}, surrogate modeling \cite{jiang2020dominant}, and deep reinforcement learning \cite{guan2022structural}.

While these approaches offer potential solutions to the issue, three significant limitations pertain. First, the first two approaches are unsuitable for unobserved datasets or for failure modes that are difficult to capture using brute-force Monte Carlo simulations (MCS). Second, while research on the third approach facilitates data generation for multiple failure modes, most studies focus only on identifying dominant failure modes, i.e., those with relatively high failure probabilities. As a result, existing research efforts remain inadequate for detecting black swan-type failure modes. Third, in addition to the second limitation, the majority of studies on the third approach have concentrated on identifying failure modes in structural systems under static loads. In other words, aleatoric uncertainties within these frameworks remain inadequately addressed.

To address these research gaps, this study proposes a framework for constructing a balanced dataset to predict failure modes of structural systems subjected to seismic excitations. The framework begins by identifying a set of critical ground motion features (GMFs) based on their correlation with structural responses. By treating GMFs as random variables alongside structural system parameters, an adaptive algorithm is proposed, inspired by the importance sampling for noteworthy scenarios (ISNS) framework \cite{kim2024efficient}. The algorithm is designed to systemically identify the probability densities of various failure domains while accounting for both epistemic uncertainties, such as material and geometric randomness, and aleatoric uncertainties arising from record-to-record variability in ground motions. The samples generated from the identified probability densities are mapped back to the ground motion time history space by using a scaling factor optimization process to construct a balanced dataset for failure mode prediction. Once the balanced dataset is constructed, we evaluate the performance of deep neural network (DNN) models in predicting failure modes under two scenarios: (i) using an imbalanced dataset typically obtained from brute-force MCS, and (ii) using a balanced dataset generated by using the proposed framework. This comparative analysis demonstrates the effectiveness of the proposed framework for predicting structural failure modes.

The remainder of the paper is organized as follows. Section~\ref{Background} defines the failure modes of structural systems and the DNN architecture developed for predicting failure modes. Section~\ref{Proposed} details the proposed framework for constructing balanced datasets, including critical GMF selection, estimation of failure modes' probability densities, and scaling factor optimization process. This section also examines the performance of DNN models using the constructed balanced dataset. Section~\ref{Examples} presents numerical investigations conducted on high-fidelity structural models to validate the applicability and effectiveness of the proposed framework. Finally, Section~\ref{Conclusion} presents the key findings and concluding remarks.

\section{Background methodologies} \label{Background}

\subsection{Failure modes of structural systems under seismic excitations} \label{FM_definition}

In earthquake engineering, the failure of a structural component $C_i$ is, in general, defined in terms of an engineering demand parameter (EDP) \cite{park1985mechanistic,applied2009quantification}. The failure event of the $i$-th component is denoted as $C_i$, while its survival is represented as $\overline{C}_i$, where $i=1,2,….,N_c$, and $N_c$ denotes the total number of structural components. Mathematically, the failure event $C_i$ is expressed as the exceedance of the EDP's capacity:
\begin{equation}  \label{Eq:Ci_failure}
C_i: \mathrm{g}_i(\mathbf{x}) = \mathrm{EDP}_{\mathrm{L},i} - \mathrm{EDP}_i(\mathbf{x}) \leq 0 \,,
\end{equation}
where $\mathrm{EDP}_{\mathrm{L},i}$ represents the seismic capacity or performance limit for the $i$-th component, $\mathrm{EDP}_i(\mathbf{x})$ denotes the seismic demand obtained from nonlinear response history analysis (RHA), and $\mathrm{g}_i(\mathbf{x})$ is the limit state function, with its negative sign indicating the occurrence of failure. Here, $\mathbf{x} = [\mathbf{x}_{\mathbf{GM}}, \mathbf{x}_{\mathbf{S}}]\in\r^{N}$ denotes the input random variables, in which $\mathbf{x}_{\mathbf{GM}}\in\r^{N_{GM}}$ captures aleatory uncertainties associated with record-to-record variability of ground motions, and $\mathbf{x}_{\mathbf{S}}\in\r^{N_{S}}$ accounts for epistemic uncertainties such as material property and geometric variability. The combined input dimension is $N=N_{GM}+N_{S}$, while the dimension of the structural responses under seismic loading, $\mathbf{EDP}\in\r^{N_c}$, is equal to the number of components $N_c$.

Based on the aforementioned definition, a set of failure modes, $\mathbf{F}$, is defined as:
\begin{equation}  \label{Eq:Fail_mode}
\mathbf{F} = \left\{F \mid F = \left(\cap_{i \in \mathbf{S}} C_i \right) \cap \left(\cap_{j \in \mathbf{S}^c} \overline{C}_j \right), \, \mathbf{S}\subset\{1,2,...,N_c\}   \right\} \,,
\end{equation}
where $F$ is an instance of failure mode, and $\mathbf{S}^c$ represents the complement of the set $\mathbf{S}$. For example, a failure mode in which only the first component fails while all others remain operational can be represented as $F = \left(C_1 \cap \overline{C}_2 \cap \cdots \cap \overline{C}_{N_c} \right)$. Considering the Boolean characteristics of structural component states and excluding the scenario of surviving all components, the total number of possible failure modes, $N_F$, can be estimated by $N_F=2^{N_c}-1$.

Identifying samples within each failure mode $F_i$ presents several challenges. Such samples are denoted as $ \mathcal{X}_{F_i} = \{ \mathbf{x} : \mathbf{x} \in \Omega(\mathbf{x} | F_i) \}$, where $\Omega(\mathbf{x} | F_i)$ represents the domain of $\mathbf{x}$ corresponding to the failure mode $F_i$. First, since failure modes are functions of random variables, some failure modes are difficult to identify due to their rarity. Second, the high dimensionality of seismic excitations ($\mathbf{x}_{\mathbf{GM}}$) hinder the direct integration of existing research efforts aimed at constructing balanced datasets \cite{kim2024dimensionality}. Moreover, due to the record-to-record variability, multiple GMFs are required to distinguish ground motion time histories, instead of relying on a single GMF such as peak ground acceleration (PGA).

\subsection{DNN model for predicting failure modes} \label{DNN}

A DNN model is designed to facilitate near-real-time prediction of structural failure modes under seismic excitations. Based on the mathematical representation of failure modes in Eq.~\eqref{Eq:Fail_mode}, the DNN model outputs the Boolean states of structural components. Each failure mode is represented in binary form, where a value of 1 indicates failure ($C_i$) and 0 indicates survival ($\overline{C}_i$). For example, a failure mode $F$=[1 0 $\cdots$ 0] in Section~\ref{FM_definition} corresponds to $\left(C_1 \cap \overline{C}_2 \cap \cdots \cap \overline{C}_{N_c} \right)$.

\begin{figure}[H]
  \centering
  \includegraphics[scale=0.37]  {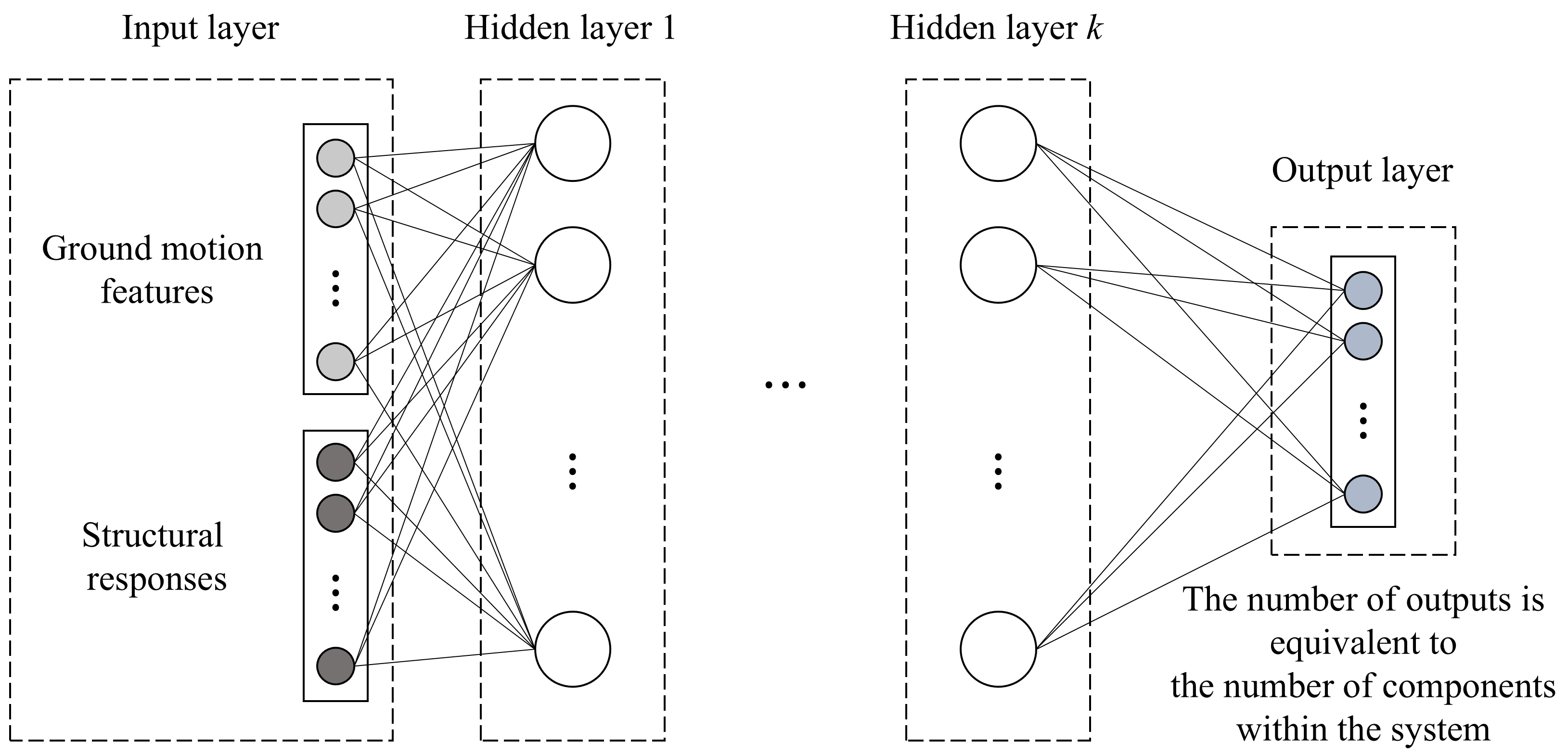}
  \caption{\textbf{Architecture of the DNN model for near-real time failure mode prediction}.}
  \label{Fig_DNN}
\end{figure}

To enable near-real-time prediction, the inputs to the model can include structural responses, such as peak acceleration value obtained from accelerometers for each building story, along with a set of GMFs. Figure~\ref{Fig_DNN} illustrates the architecture of the proposed DNN model. Note that a separate DNN model is constructed for each structural system under investigation.

The Rectified Linear Unit (ReLU) is utilized as the activation function for the hidden layers, while a sigmoid function is employed in the final hidden layer to predict whether a component has failed or not. The number of hidden layers and units varies based on numerical investigations, with details provided in each subsection. The number of units in the output layer corresponds to the number of structural components under consideration, i.e., $N_c$. Section~\ref{Proposed} presents a framework for constructing a balanced dataset for training the DNN model. Note that since the DNN architecture is designed primarily for demonstrating the effectiveness of the constructed balanced dataset, alternative configurations and input types can also be employed.

\section{Construction of balanced datasets for predicting failure modes} \label{Proposed}

\begin{figure}[H]
  \centering
  \includegraphics[scale=0.56] {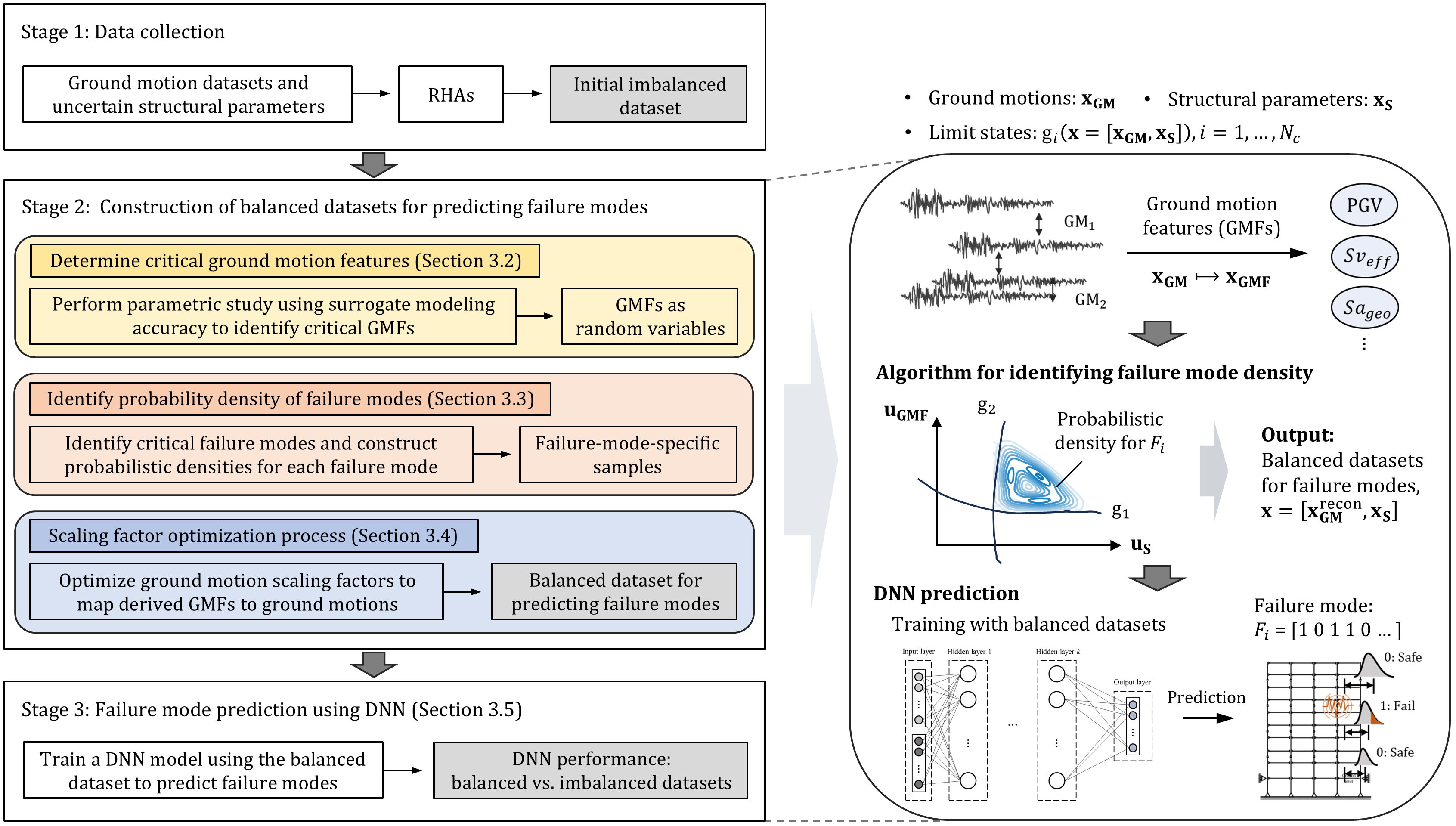}
  \caption{\textbf{Proposed framework for constructing balanced datasets for failure mode prediction}.}
  \label{Fig_overview}
\end{figure}

Figure~\ref{Fig_overview} provides a comprehensive overview of the entire procedure from data preprocessing to the demonstration of the proposed algorithm using DNN models. Section~\ref{GM_and_MDOF} presents a ground motion dataset and a benchmark structural model used as an illustrative example in this study. In other words, we develop a balanced seismic demand database using selected ground motions in Section~\ref{GM_and_MDOF}. Section~\ref{GMF_selection} details the selection of critical GMFs that effectively represent the variability of structural responses. Sections~\ref{FailMode_Algo} and \ref{Construction} describe the identification of the probability densities of critical failure mode domains and the scaling factor optimization process, respectively. The scaling factor optimization process is designed to identify the appropriate ground motion and scaling factor that align with the critical GMFs of the generated samples derived from the identified probability density function. This selected ground motion is then used in nonlinear RHAs for generating structural responses. The collection of these responses forms a balanced dataset. To demonstrate the efficacy of the proposed algorithm, Section~\ref{DNN_results} presents numerical investigations that evaluate the performance of the DNN model trained on (i) an imbalanced dataset obtained from MCS and (ii) a balanced dataset generated using the proposed framework.

\subsection{Ground motion dataset and benchmark structural model} \label{GM_and_MDOF} 

\begin{figure}[H]
  \centering
  \includegraphics[scale=0.53] {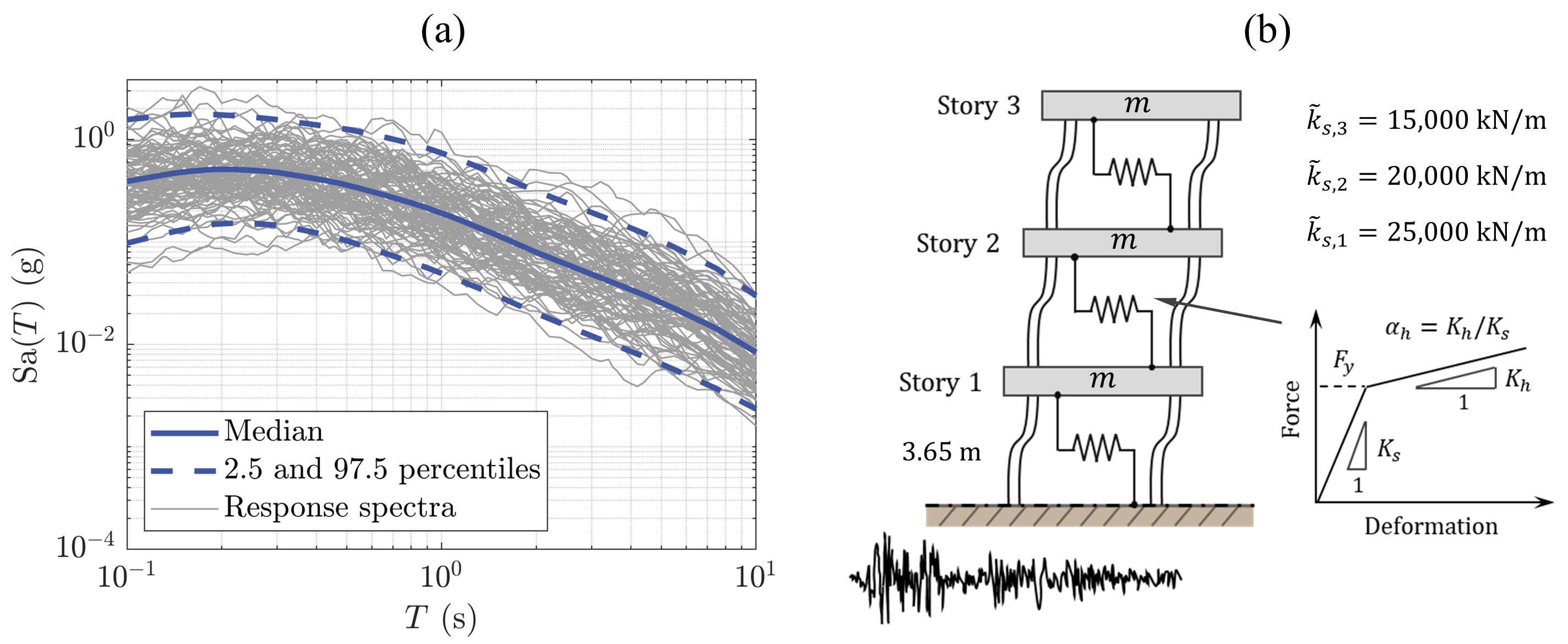}
  \caption{\textbf{(a) Response spectra of selected ground motions with the target spectrum and (b) structural model}. Figure (a) shows 100 response spectra with the median and 2.5\%-97.5\% quantiles of the target spectrum superimposed.}
  \label{Fig_MDOF_spectra}
\end{figure}

According to modern seismic design standards, a set of spectrum-compatible ground motions is required for the design and performance assessment of structural systems \cite{asce2010minimum}. To achieve this, a set of ground motions is selected to be compatible with a spectrum derived from the ground motion prediction equation (GMPE) developed by Boore and Stewart \cite{boore2014nga} and the spectral correlation model proposed by Baker and Jayaram \cite{baker2008correlation}. Table~\ref{Tab_TS} summarizes the seismic hazard parameters, representing a moderate seismic scenario. A ground motion selection method \cite{baker2018improved} is employed to identify recorded motions that align with the target spectrum. This method selects ground motions with minimal errors relative to the target spectrum and applies scaling for improved spectral alignment. As a result, 523 ground motions are selected from the PEER NGA-West 2 database \cite{ancheta2014nga}. Figure~\ref{Fig_MDOF_spectra}(a) presents the response spectra of the selected ground motions, along with the median and 2.5\%–97.5\% quantiles of the target spectrum.

\begin{table}[H]
  \caption{\textbf{Seismic hazard parameters defining the target spectrum}.}
  \label{Tab_TS}
  \centering
  \begin{tabular}{l l}
    \toprule
    Parameter & Value \\
    \midrule
    Earthquake magnitude & 6.8 \\
    Closest distance to fault rupture (km) & 15 \\
    Average shear wave velocity (m/s) & 450 \\
    Fault type & Strike-slip \\
    Region & California \\
    \bottomrule
  \end{tabular}
\end{table}

To illustrate the proposed framework, a three-story shear-type multi-degree-of-freedom (MDOF) building shown in Figure~\ref{Fig_MDOF_spectra}(b) is modeled using OpenSees \cite{mckenna2011opensees}, incorporating bilinear springs to capture nonlinearities. Yield force ($F_y$) and strength hardening ratio ($\alpha_h$) are treated as deterministic, defined as $F_y=10 \,\text{mm} \times K_s$ and $\alpha_h=4.5\%$, respectively, whereas damping ratio ($\xi$), floor mass ($m$), and story stiffness ($K_s$) are treated as random variables. In other words, epistemic uncertainties are represented as $\mathbf{x}_{\mathbf{S}}=[\xi,m,u_k]$. The damping ratio and floor mass follow lognormal distributions with means of 0.03 and 90,000 kg, respectively, and coefficients of variation (CoV) of 0.25. The story stiffness is modeled as $K_s=\tilde{k}_s e^{\sigma_k u_k}$, where $\sigma_k=0.25$, $u_k$ represents the standard Gaussian variable, and $\tilde{k}_s$ denotes the nominal stiffness values of 25,000 kN/m, 20,000 kN/m, and 15,000 kN/m for the first, second, and third stories, respectively.

\begin{figure}[H]
  \centering
  \includegraphics[scale=0.58] {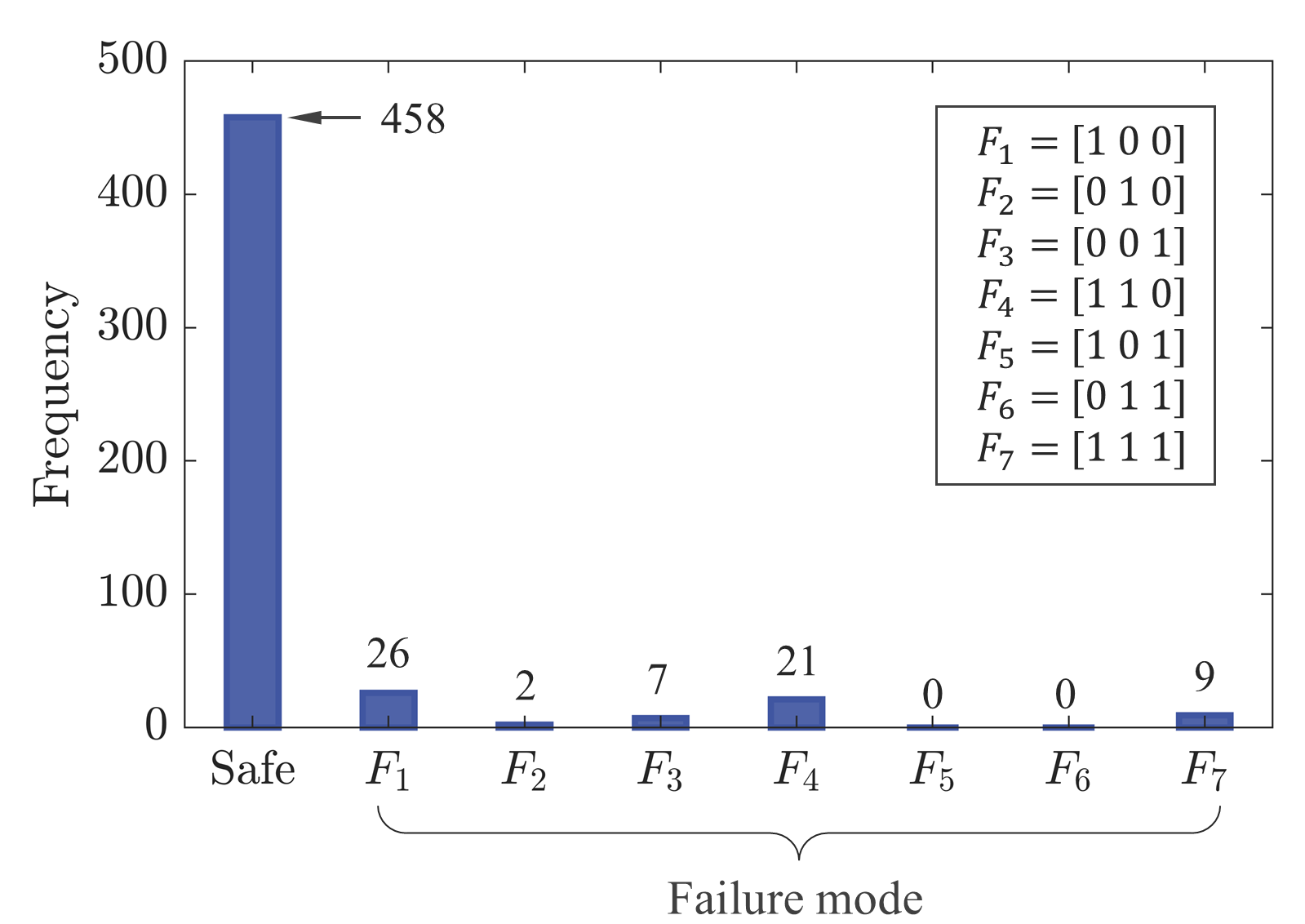}
  \caption{\textbf{Histogram of failure modes obtained through MCS}. This represents an \textit{imbalanced} dataset dominated by safe scenario.}
  \label{Fig_Failure_mode_MCS}
\end{figure}

By setting the peak inter-story drift ratio (IDR) as the EDP of interest, MCS is performed to construct a seismic demand database. Failure is defined as an IDR exceeding 1.7\%. Figure~\ref{Fig_Failure_mode_MCS} presents the distribution of failure modes obtained from 523 simulations. Note that each selected ground motion is used only once for RHA. During RHA with these ground motions, the structural parameters $\mathbf{x}_{\mathbf{S}}$ are randomly generated according to the statistical distribution defined in the previous paragraph. However, ground motions could be used multiple times with different structural parameter realizations or by varying the ground motion scaling factor.

Among $2^3-1=7$ possible failure modes, $F_1$ corresponds to the failure mode [1 0 0], where only the first story fails, while $F_7$ represents the mode [1 1 1], where all stories fail. The histogram reveals a predominance of the safe scenario, with limited samples in failure domains, emphasizing the need to construct balanced datasets to address this imbalance.

\subsection{Identification of critical ground motion features} \label{GMF_selection} 

To effectively incorporate the aleatoric uncertainty, a set of critical GMFs needs to be defined. These features, denoted as $\mathbf{x}_{\mathbf{GMF}}$, are selected based on their ability to capture the essential attributes of ground motions that influence structural responses. The selection process employs Gaussian process (GP) models to evaluate the relationship between the feature space, $\mathbf{x}'=[\mathbf{x}_{\mathbf{GMF}}, \mathbf{x}_{\mathbf{S}}]$, and the EDPs. The introduction of GP model is to ensure consistency with the proposed GP-based adaptive algorithm for identifying failure mode density, as detailed in subsequent sections.

Table~\ref{Tab_GMfeatures} summarizes a total of 18 candidate GMFs from the literature \cite{padgett2008selection,kim2021clustering,kim2024deep}. A damping ratio of 5\% is assumed when calculating spectral values. Among the candidate GMFs, critical GMFs are identified through cross-validation using the structural model and the set of ground motions introduced in Section~\ref{GM_and_MDOF}. To this end, the 523 samples generated in Section~\ref{GM_and_MDOF} are first divided into training subset of 100 samples and test subset of 423 samples. GP models are then trained to predict EDPs for different number of GMFs, ranging from individual features to combinations of up to 16 features. Note that for a given input feature set, three GP models are trained to predict IDR for each story. The performance of the GP models is evaluated using R-squared values ($R^2$) for EDP predictions on the test set within the feature space.
\begin{table}[h]
  \caption{\textbf{Summary of candidate ground motion features (GMFs) considered for feature selection}.}   \label{Tab_GMfeatures}
  \centering
  \resizebox{\textwidth}{!}{ 
  \renewcommand{\arraystretch}{1.25} 
  \begin{tabular}{c c l}
    \toprule
    Feature & Unit & Description \\
    \midrule
    PGA & m/s$^2$ & \\
    PGV & m/s &  \\
    PGD & m & \multirow{-3}{43em}{Peak ground acceleration (PGA), velocity (PGV), and displacement (PGD).} \\ 
    \rowcolor{gray!10} $Sa(T_1)$ & m/s$^2$ &  \\
    \rowcolor{gray!10} $Sv(T_1)$ & m/s &  \\
    \rowcolor{gray!10} $Sd(T_1)$ & m & \multirow{-3}{43em}{Maximum pseudo-acceleration ($Sa$), velocity ($Sv$), and displacement ($Sd$) for a linear oscillator with the first mode period $T_1$. Note that $Sv(T_1) \approx Sa(T_1)/\omega$ and $Sd(T_1) \approx Sa(T_1)/\omega^2$, where $\omega$ denotes circular frequency.}  \\
    $Sa_{geo}$ & m/s$^2$ & \\
    $Sv_{geo}$ & m/s &  \\
    $Sd_{geo}$ & m & \multirow{-3}{43em}{Geometric mean of pseudo-acceleration, velocity, and displacement over periods ranging from 0.1 to 2.5 seconds. For example, $Sa_{geo}$ is defined as $\left( \prod_{i=1}^{n} Sa(T_i) \right)^{1/n}$.} \\
    \rowcolor{gray!10} $Sa_{eff}$ & m/s &  \\
    \rowcolor{gray!10} $Sv_{eff}$ & m &  \\
    \rowcolor{gray!10} $Sd_{eff}$ & m$\cdot$s & \multirow{-3}{43em}{Effective spectral acceleration, velocity, and displacement integrated over periods ranging from 0.1 to 2.5 seconds. For example, $Sv_{eff}$ is defined as $\int_{0.1}^{2.5} Sv(T)\,dT$, also referred to as Housner intensity \cite{housner1965intensity}.} \\
    PGV/PGA & s & Ratio of PGV to PGA. \\
    \rowcolor{gray!10} \begin{tabular}{@{}c@{}} Spectral\\ shape \end{tabular} & m/s$^2$ & \begin{tabular}{@{}l@{}} Geometric mean of $Sa(T_1)$ and $Sa(2\cdot T_1)$, reflecting nonlinear structural behavior\\ such as period elongation \cite{cordova2000development}. \end{tabular} \\
    $I_A$ & m/s & \begin{tabular}{@{}l@{}} Arias intensity, defined as the integral of squared ground motion acceleration\\ over the motion duration $T_{d}$: $I_A = \frac{\pi}{2g}\int_{0}^{T_{d}} a(t)^2\,dt$ \cite{arias1970measure}. \end{tabular} \\
    \rowcolor{gray!10} $D_{5-95}$ & s & Duration between 5\% and 95\% of Arias intensity, indicative of shaking duration. \\
    \bottomrule
  \end{tabular} 
  } \\
\end{table}

\begin{figure}[H]
  \centering
  \includegraphics[scale=0.48] {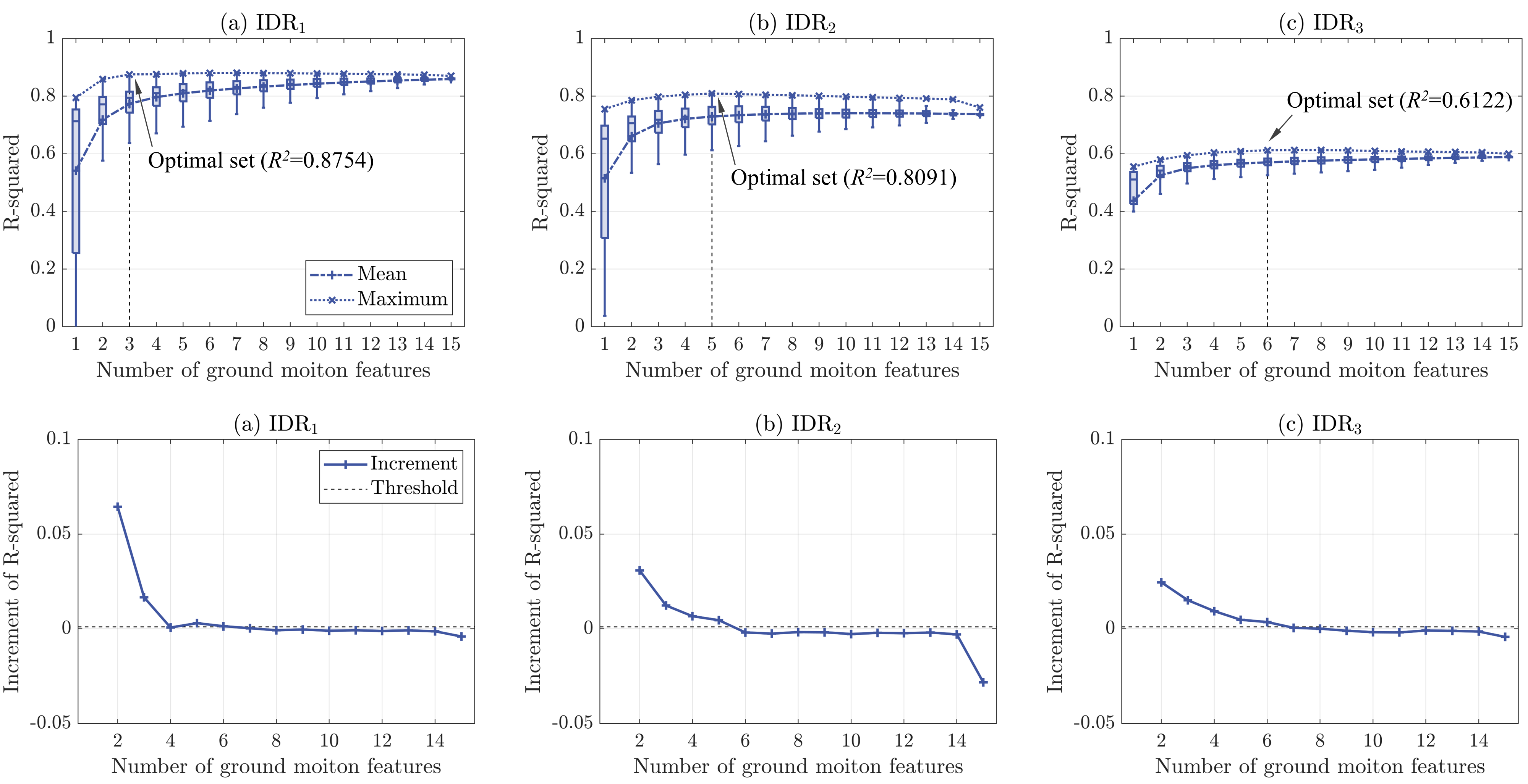}
  \caption{\textbf{R-squared values (first row) and incremental improvements (second row) for different GMF combinations: (a) IDR$_1$, (b) IDR$_2$, and (c) IDR$_3$}.}
  \label{Fig_GMfeature_R2}
\end{figure}

The first row of Figure~\ref{Fig_GMfeature_R2} illustrates $R^2$ values for various GMF combinations corresponding to three EDPs: IDR at the 1st, 2nd, and 3rd stories. The x-axis represents the number of GMFs, and each box plot summarizes the results for all possible feature combinations at that number. For instance, for a GMF set having seven features, $\binom{16}{7} = 11,440$ modeling results are summarized in a box plot. In addition to the $R^2$ value, the incremental improvement in $R^2$ between successive steps, calculated as $\Delta_{R^2} = R^2_i-R^2_{i-1}$, is shown in the second row, where $i$ represents the number of GMFs starting from 2.

As the number of GMFs increases, the improvement in $R^2$ generally decreases, indicating that a representative subset of GMFs is adequate to capture the variability in seismic responses. Negative increments in $R^2$ observed in some cases may indicate potential overfitting when excessive GMFs are included. The optimal number of GMFs for identifying the critical GMFs is selected based on a threshold of $\Delta_{R^2}<0.001$, ensuring adequate representation while avoiding overfitting. Among the combinations satisfying this criterion, the one maximizing $R^2$ is selected as the optimal GMF set, highlighted with dashed lines in the first row of Figure~\ref{Fig_GMfeature_R2} and listed in Table~\ref{Tab_GMfeatures_identified}.

\begin{figure}[H]
  \centering
  \includegraphics[scale=0.60] {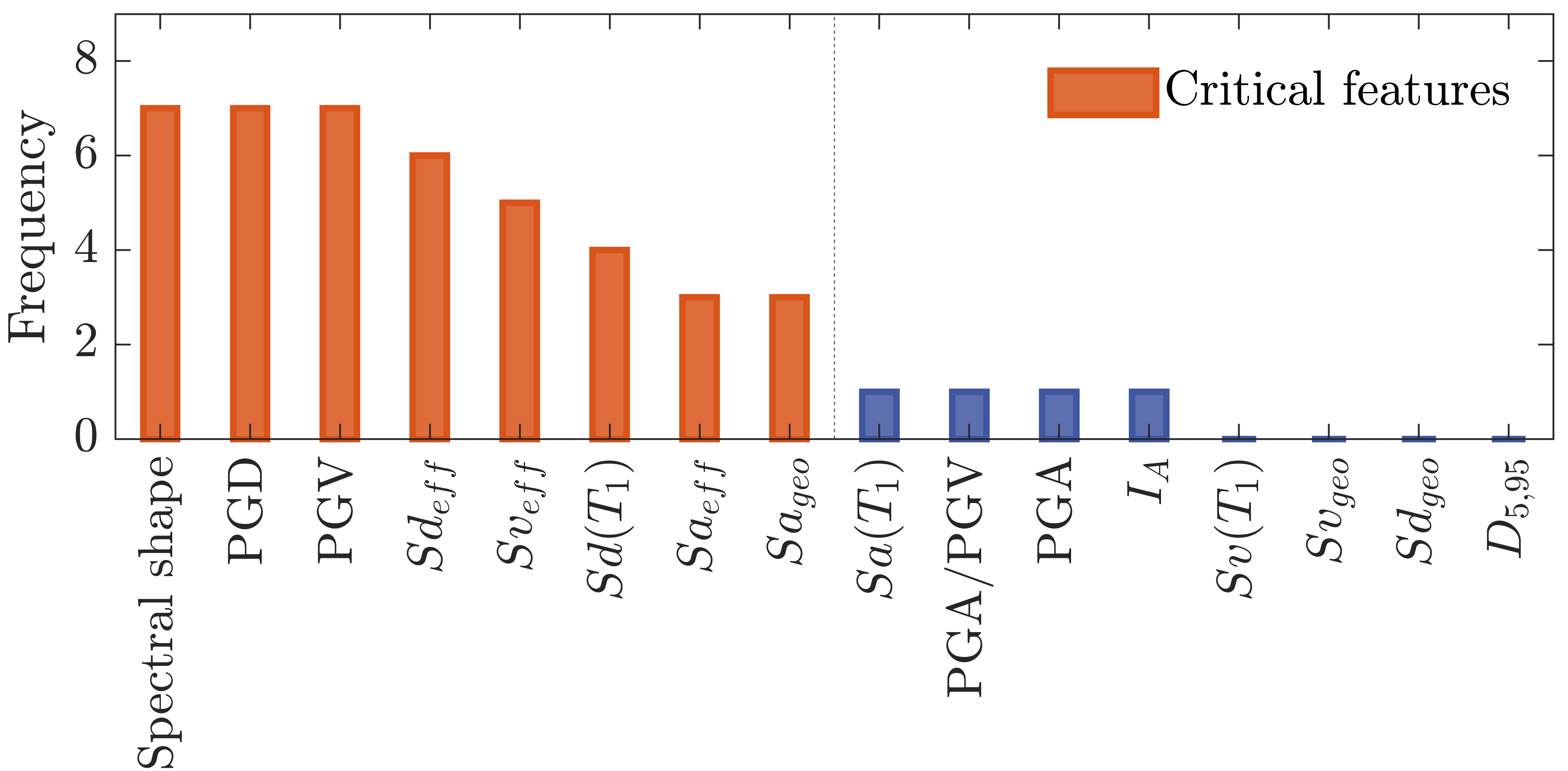}
  \caption{\textbf{Frequencies of GMFs identified as critical}. This histogram summarizes the GMFs selected across multiple training set sizes, highlighting the eight features chosen for the proposed framework.}
  \label{Fig_GMfeature_final}
\end{figure}

To examine the impact of training dataset size on the identification of critical GMFs, we repeat the identification process using training set sizes of 200 and 300 samples, respectively. The same procedure described earlier is applied. Table~\ref{Tab_GMfeatures_identified} lists the identified GMFs for each EDP under training set sizes of 200 and 300 samples in addition to the 100 samples. While some variations in critical GMFs are observed, similar GMFs are consistently selected across different training set sizes. Based on these results, Figure~\ref{Fig_GMfeature_final} summarizes the frequencies of GMFs identified as critical, showing that eight features are consistently identified: spectral shape, PGV, PGD, $Sd_{eff}$, $Sv_{eff}$, $Sd(T_1)$, $Sa_{eff}$, and $Sa_{geo}$. These GMFs form $\mathbf{x}_{\mathbf{GMF}}$, effectively representing ground motion variability in the proposed framework. These selections align well with previous literature \cite{padgett2008selection,kim2021clustering,hu2023unsupervised,kim2024deep,ding2024feature}, which analyzed the effect of ground motion on the peak response of structures.

\begin{table}[h]
  \caption{\textbf{Critical GMFs identified for each EDP across different training set sizes}.}   \label{Tab_GMfeatures_identified}
  \centering
  \resizebox{0.95\textwidth}{!}{ 
  \renewcommand{\arraystretch}{1.02} 
  \begin{tabular}{c c c c c c c c c}
    \toprule
    \begin{tabular}{@{}c@{}} Training\\set size \end{tabular} & EDP & \begin{tabular}{@{}c@{}} Number of\\GMFs \end{tabular} & \multicolumn{6}{c}{Identified GMFs}  \\
    \midrule
    \multirow{3}{*}{100} & IDR$_1$ & 3 & PGV & Spectral shape & $Sv_{eff}$ &  &  &  \\
     & IDR$_2$ & 5 & PGV & PGD & $Sa(T_1)$ & Spectral shape & $Sd_{eff}$ &  \\
     & IDR$_3$ & 6 & PGV & PGD & Spectral shape & $Sa_{geo}$ & $Sv_{eff}$ & $Sd_{eff}$ \\ 
    \midrule
    \multirow{3}{*}{200} & IDR$_1$ & 6 & PGV & PGD & Spectral shape & $Sv_{eff}$ & $Sd_{eff}$ & $Sa_{geo}$ \\
     & IDR$_2$ & 5 & PGV & PGD & $Sd(T_1)$ & $Sa_{eff}$ & $Sd_{eff}$ &  \\
     & IDR$_3$ & 5 & PGD & PGV/PGA & Spectral shape & $Sa_{eff}$ & $I_A$ & \\ 
    \midrule
    \multirow{3}{*}{300} & IDR$_1$ & 4 & PGV & $Sd(T_1)$ & $Sv_{eff}$ & $Sd_{eff}$ & & \\
     & IDR$_2$ & 6 & PGV & PGD & $Sd(T_1)$ & Spectral shape & $Sa_{geo}$ & $Sv_{eff}$ \\
     & IDR$_3$ & 6 & PGA & PGD & Spectral shape & $Sd(T_1)$ & $Sa_{eff}$ & $Sd_{eff}$ \\ 
    \bottomrule
  \end{tabular}
  } \\
\end{table}

Based on comprehensive numerical investigations and a thorough literature review, this study identifies eight critical GMFs using a three-story MDOF system and a specific ground motion set. These GMFs are deemed sufficient to capture the variability in seismic responses for commonly configured structural systems subjected to ground motion sets compatible with a general design spectrum. Consequently, the identified GMFs are utilized throughout this manuscript. Nevertheless, if the proposed framework is applied to complex structural systems with unique ground motion sets, it is required to identify new critical GMFs following the procedure outlined in this section.

\subsection{Algorithm for identifying probability density of failure modes} \label{FailMode_Algo}

Inspired by the ISNS algorithm \cite{kim2024efficient}, a failure mode identification framework is proposed. The failure mode identification procedure, illustrated in Figure~\ref{Fig_FailMode}, proceeds through the following steps:
\begin{figure}[H]
  \centering
  \includegraphics[scale=0.65] {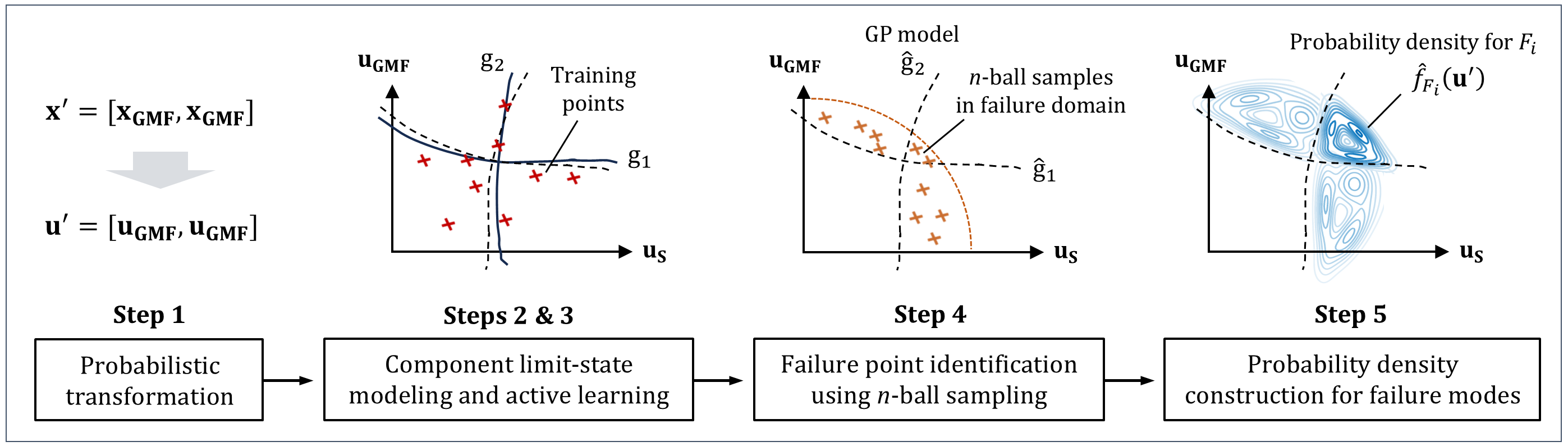}
  \caption{\textbf{Illustration of the algorithm for identifying the probability density of failure modes}.}
  \label{Fig_FailMode}
\end{figure}
\begin{enumerate}
    \item \textbf{Probabilistic transformation}: Input variables are transformed into the standard Gaussian space using the Nataf model \cite{der2022structural}, leading to uncorrelated standard Gaussian variables, $\mathbf{u}' = [\mathbf{u}_{\mathbf{GMF}}, \mathbf{u}_{\mathbf{S}}]$. This transformation ensures that GMFs and structural parameters are represented in a space with rotational symmetry, facilitating efficient exploration of failure domains. GMFs are transformed using empirical cumulative distribution functions (CDFs) derived from the dataset of 523 ground motions introduced in Section~\ref{GM_and_MDOF}. These transformed variables serve as the input space for subsequent analyses.
    
    \item \textbf{Component limit state modeling}: Using a subset of the total samples in the pervious step (e.g., 30\%), GP surrogate models, $\hat{\mathrm{g}}_k(\mathbf{u}')$, are constructed for each structural component. The role of the surrogate models is to approximate their respective limit state functions. These models enable efficient identification of failure domains and are integrated with $n$-ball sampling. Details of the of GP modeling are provided in~\ref{App:GP}.
    
    \item \textbf{Adaptive learning}: An active learning strategy is implemented to iteratively update the GP models by selecting training points that maximize information gain. In other words,  new points are chosen from the total sample set, combined with the previously used training points, and subsequently utilized to retrain the GP models. This approach reduces redundant RHAs and enhances computational efficiency. The optimal training point, $\mathbf{u}'_*$, is identified by minimizing a composite learning criterion:
    \begin{equation}  \label{Eq:LF}
    \mathbf{u}'_* = \mathop{\arg\min}_{\mathbf{u}'} \alpha(\mathbf{u}') = \mathop{\arg\min}_{\mathbf{u}'} \frac{ \left| \mu_{\hat{\mathrm{g}}_{ct}}(\mathbf{u}') \right| }{\sigma_{\hat{\mathrm{g}}_{ct}}(\mathbf{u}')} \,,
    \end{equation}
    where $\mu_{\hat{\mathrm{g}}_{ct}}(\mathbf{u}')$ and $\sigma_{\hat{\mathrm{g}}_{ct}}(\mathbf{u}')$ are respectively the GP mean and standard deviation of the most critical limit state function (indexed by $ct$), defined as the function with the smallest $\left| \mu_{\hat{\mathrm{g}}_{i}}(\mathbf{u}') \right|$ for $i=1,...,N_c$. This criterion consolidates multiple components into a single learning function. The training dataset is iteratively updated with RHA evaluations at $\mathbf{u}'_*$ until the convergence criterion, $\min \alpha(\mathbf{u}') > 2$, is met.
    
    \item \textbf{Failure point identification using $n$-ball sampling}: Uniform samples are generated within an $n$-dimensional hypersphere of radius $R$, where $n$ represents the dimension of $\mathbf{u}'$. The joint probability density function (PDF) of the $n$-ball samples is defined as $h_b(\mathbf{u}')=1/{V_b(R)}$ when $\|\mathbf{u}'\|_2 \leq R$, and $h_b(\mathbf{u}')=0$ otherwise, where $V_b(R)$ denotes the volume of the hypersphere. Within the $n$-ball domain, failure points with occurrence probabilities below $\Phi(-R)$, where $\Phi$ is the standard Gaussian CDF, are identified. In other words, the $n$-ball samples, denoted as $\mathcal{U}_{b}=\{\mathbf{u}'_i,i=1,...,n_{b} \}$, are evaluated using GP models to identify failure points associated with each failure mode. The identified failure points for the $i$-th failure mode are expressed as:
    \begin{equation}  \label{Eq:FailPoints}
    \mathcal{U}_{F_i} = \{ \mathbf{u}' \in \mathcal{U}_{b} : \mathbf{u}' \in \hat{\Omega}(\mathbf{u}' | F_i) \} \,,
    \end{equation}
    where $\hat{\Omega}(\mathbf{u}' | F_i)$ represents the GP-predicted failure domain for mode $F_i$ (defined by replacing $g$ with $\hat{\mathrm{g}}$ in Eqs.~\eqref{Eq:Ci_failure} and~\eqref{Eq:Fail_mode}. Note that the radius of hypersphere $R$ is defined based on the decision-maker's preference to establish trivial cases of failure modes.

    \item \textbf{Probability density construction for failure modes}: The probability density of each failure mode is approximated using a Gaussian mixture model \cite{reynolds2009gaussian} fitted to the $n$-ball samples within the identified failure regions, $\mathcal{U}_{F_i}$. For the $i$-th failure mode, the density describing the samples $\mathcal{U}_{F_i}$ is expressed as:
    \begin{equation} \label{Eq:ISNS_qi}
    \hat{f}_{F_i}(\mathbf{u}') = \sum_{k=1}^{n_m} \varphi^i_k f_N(\mathbf{u}'; \mathbf{\mu}_k^i; \mathbf{\Sigma}_k^i)   \,,
    \end{equation}
    where $f_N$ is the Gaussian PDF, $n_m$ is the number of mixture components, and $\varphi^i_k$, $\mathbf{\mu}_k^i$, and $\mathbf{\Sigma}_k^i$ are the mixture weights, mean vectors, and covariance matrices, respectively, determined using the expectation-maximization algorithm.   
\end{enumerate}

\begin{figure}[H]
  \centering
  \includegraphics[scale=0.50] {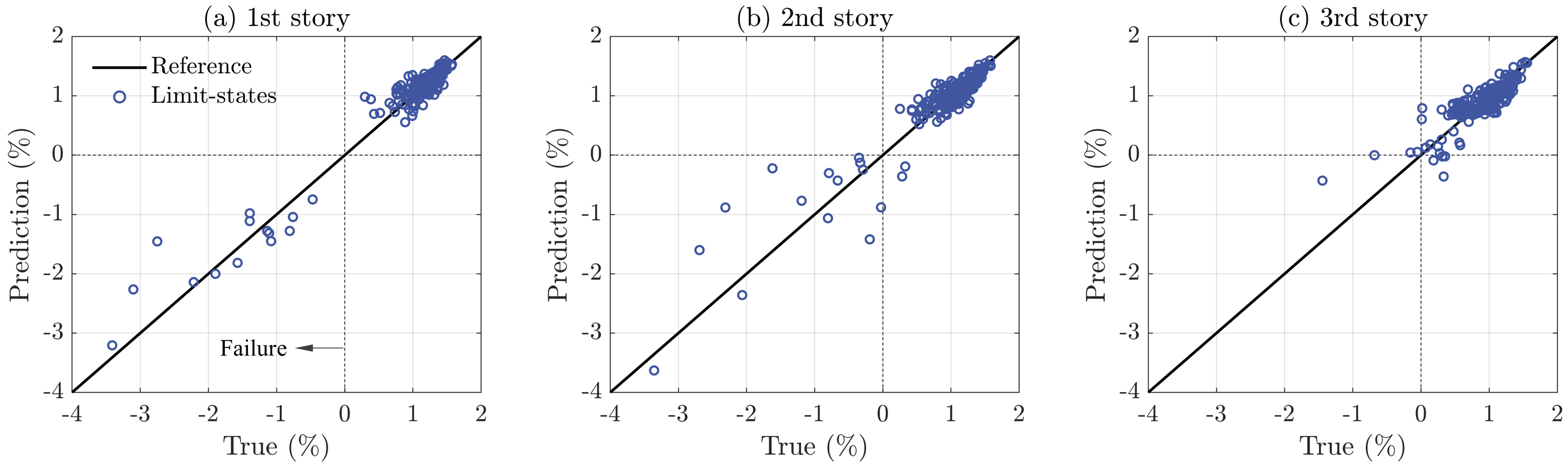}
  \caption{\textbf{Scatter plots of true limit state values against GP predictions for the MDOF building structure: (a) 1st story, (b) 2nd story, and (c) 3rd story}. The solid line represents perfect agreement between predictions and observations, while the dashed line indicates the failure threshold.}
  \label{Fig_GP_scatter}
\end{figure}

Starting from 200 initial training points, the adaptive algorithm selects 97 more samples to construction GP models, i.e., results up to Step 3 in the framework. Figure~\ref{Fig_GP_scatter} presents scatter plots of the three limit states, $\mathrm{g}_i$, against their GP predictions, $\hat{\mathrm{g}}_i$, constructed using 297 training samples. The trained GP models exhibit high predictive performance, with $R^2$ values of 0.9535, 0.8630, and 0.7301 for the 1st, 2nd, and 3rd limit states, respectively.

Utilizing $n_{b}=10^7$ samples generated within an $n$-ball of radius $R = 6.0$ and $n_m = 3$, the framework identifies five critical failure modes: $F_1$, $F_2$, $F_3$, $F_4$, $F_7$, as summarized in Table~\ref{Tab_FM_MDOF}. For simplicity, the cap notation is omitted in the third column of the table. The probability densities of these modes serve as the sampling distributions for balanced dataset construction. Note that failure modes with occurrence probabilities below $\Phi(-6.0)=9.8\times 10^{-10}$ are considered trivial and excluded from the analysis. To explore failure modes with extremely low probabilities, one can increase the radius of the $n$-ball sampling and enhance the variability of the sample set in Step 1.

\begin{table}[H]
  \caption{\textbf{Failure modes identified for the MDOF building structure}.}
  \label{Tab_FM_MDOF}
  \centering
  \begin{tabular}{c c c}
    \toprule
    Case & Mode index ($i$) & Failure mode ($F_i$) \\
    \midrule
    1 & 1 & $\left\{C_1 \overline{C}_2 \overline{C}_{3} \right\}$ \\
    2 & 2 & $\left\{\overline{C}_1 C_2 \overline{C}_{3} \right\}$ \\
    3 & 3 & $\left\{\overline{C}_1 \overline{C}_2 C_{3} \right\}$ \\
    4 & 4 & $\left\{C_1 C_2 \overline{C}_{3} \right\}$ \\
    5 & 7 & $\left\{C_1 C_2 C_{3} \right\}$ \\
    \bottomrule
  \end{tabular}
\end{table}

\subsection{Balanced dataset construction via scaling factor optimization} \label{Construction}

Once the failure-mode-specific samples are obtained in the standard Gaussian space, they must be transformed into the domain of ground motion time histories to construct the balanced dataset. This task is important as the generated samples which follow the probability density shown in Eq~\eqref{Eq:ISNS_qi}, $\mathbf{x}' = [\mathbf{x}_{\mathbf{GMF}}, \mathbf{x}_{\mathbf{S}}]$ do not directly correspond to the ground motion time histories $\mathbf{x} = [\mathbf{x}_{\mathbf{GM}}, \mathbf{x}_{\mathbf{S}}]$.

To address this issue, a scaling factor optimization process is proposed. Scaling is a widely accepted practice in earthquake engineering, enabling the modification of ground motions to achieve desired characteristics from a finite dataset \cite{padgett2008selection,baker2018improved,ding2024feature}. The objective of this process is to determine an optimal scaling factor and corresponding ground motion that minimize the discrepancy between the GMFs derived from the scaled ground motion and the generated sample $\mathbf{x}' = [\mathbf{x}_{\mathbf{GMF}}, \mathbf{x}_{\mathbf{S}}]$. The mismatch is quantified using the root mean squared error (RMSE), defined as:
\begin{equation} \label{Eq:scale_error}
\varepsilon^{j}(\gamma) = \text{RMSE} \left( \mathbf{x}_{\mathbf{GMF}}^{j}(\gamma), \mathbf{x}_{\mathbf{GMF}} \right)  \,,
\end{equation}
where $j\in\{1,...,n_{data}\}$ indexes the ground motions in the dataset ($n_{data}=523$ in this study), $\gamma$ is the scaling factor, and $\mathbf{x}^{j}_{\mathbf{GM}}(\gamma)$ represents the GMFs derived from the $j$-th ground motion scaled by factor $\gamma$.

\begin{figure}[H]
  \centering
  \includegraphics[scale=0.50] {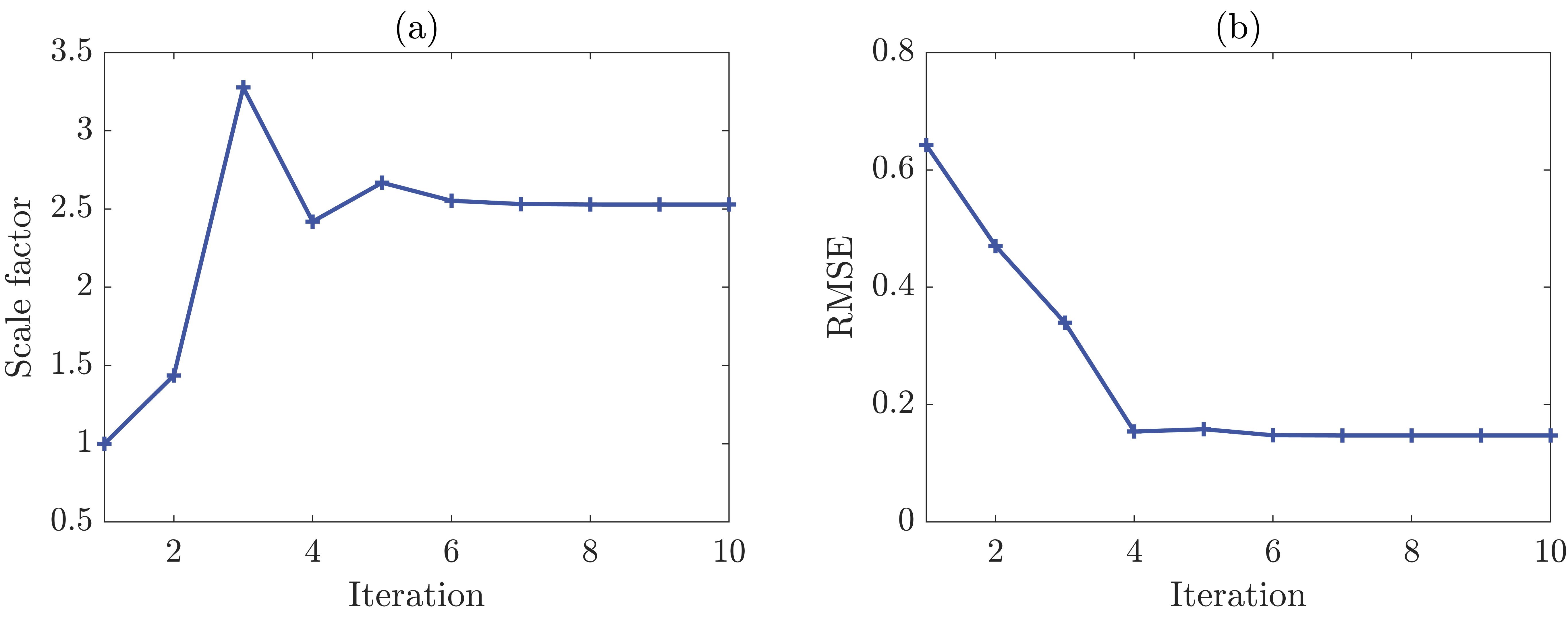}
  \caption{\textbf{Convergence histories of (a) scaling factor and (b) RMSE during optimization}.}
  \label{Fig_Scale_optim}
\end{figure}

The optimal scaling factor for each ground motion, $\gamma^{j}_*$, is obtained by solving:
\begin{equation} \label{Eq:scale_optim1}
\gamma^{j}_* = \mathop{\arg\min}_{\gamma} \varepsilon^{j}(\gamma) \,.
\end{equation}
Figure~\ref{Fig_Scale_optim} illustrates an instance of the convergence history of the scaling factor optimization. Figure~\ref{Fig_Scale_optim}(a) presents the evolution of the scaling factor over iterations, while Figure~\ref{Fig_Scale_optim}(b) depicts the corresponding convergence history of the RMSE. In this case, an optimal scaling factor of $\gamma^{j}_* = 2.529$ is identified, achieving a minimal RMSE of 0.147.

This optimization is performed for all $n_{data}$ ground motions in the dataset. Among these, the ground motion corresponding to the overall minimum error is selected as:   
\begin{equation} \label{Eq:scale_optim2}
j_* = \mathop{\arg\min}_{j=1,\ldots,n_{data}} \varepsilon^{j}(\gamma_*^{j}) \,.
\end{equation}
The reconstructed ground motion corresponding to the selected index is then expressed as: 
\begin{equation} \label{Eq:XGM_recon}
\mathbf{x}_{\mathbf{GM}}^{\mathrm{recon}} = \mathbf{x}^{j_*}_{\mathbf{GM}}(\gamma^{j_*}_*) \,.
\end{equation}
This process ensures that each generated sample $\mathbf{x}'$ is associated with a physically meaningful reconstructed ground motion. By iterating over all generated samples, the framework constructs a balanced dataset in the original parameter space. The reconstructed ground motions, combined with their associated structural parameters, form the final dataset as $\mathbf{x} = [\mathbf{x}_{\mathbf{GM}}^{\mathrm{recon}}, \mathbf{x}_{\mathbf{S}}]$. Note that in this process, the maximum scaling factor is limited to 7 to prevent excessive distortion of the original ground motion characteristics.

\begin{figure}[H]
  \centering
  \includegraphics[scale=0.58] {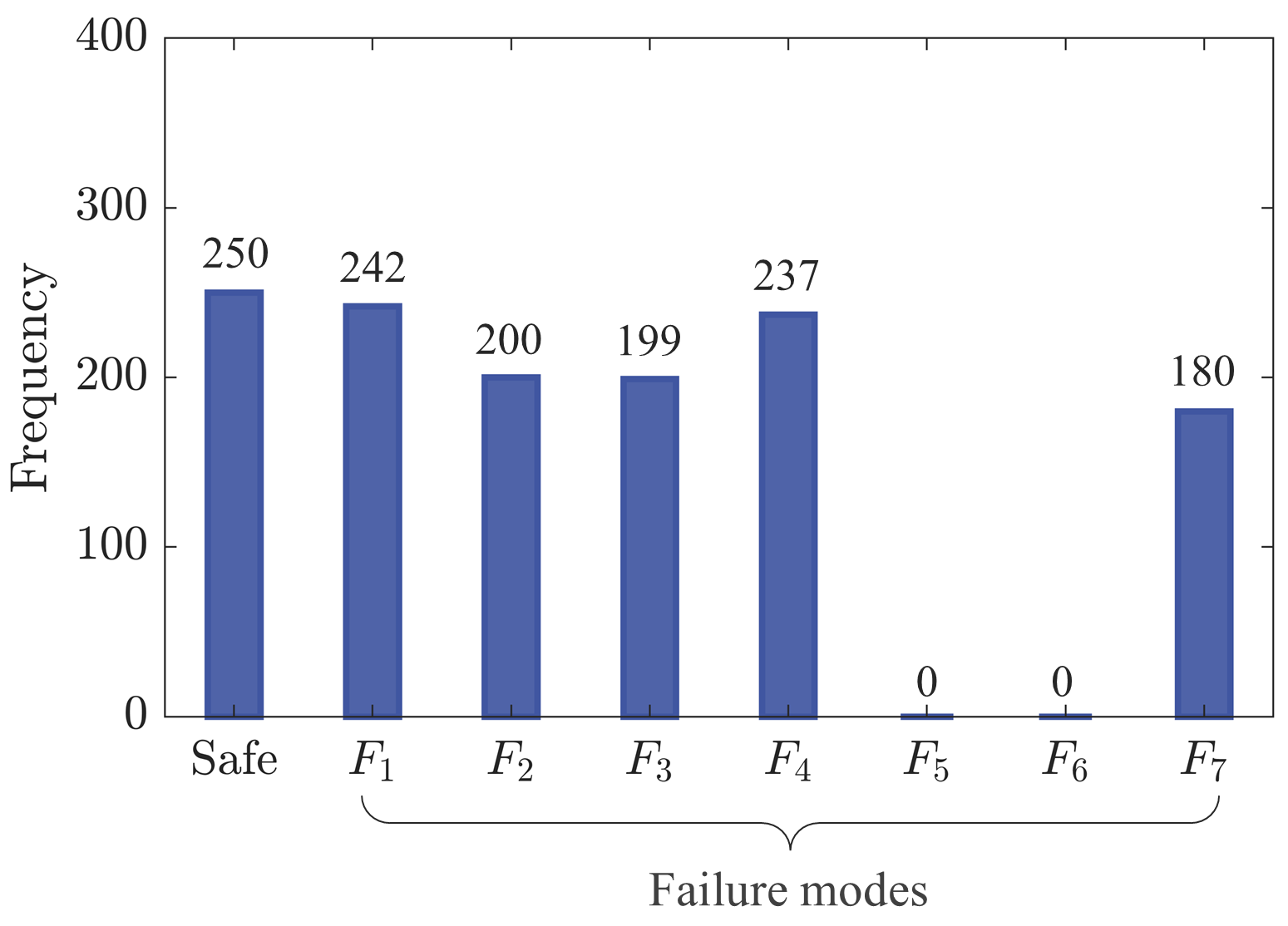}
  \caption{\textbf{Histogram of failure modes obtained using the proposed framework}. This represents a \textit{balanced} dataset that ensures sufficient representation across identified critical failure modes.}
  \label{Fig_Failure_mode_ISNS}
\end{figure}

RHAs are performed on this reconstructed dataset to evaluate the distribution of failure modes. Figure~\ref{Fig_Failure_mode_ISNS} illustrates the reconstructed dataset, which appropriately captures the identified failure modes reported in Table~\ref{Tab_FM_MDOF}. Note that inherent approximation errors arising from the reconstruction of ground motions during scaling factor optimization lead to minor deviations in the sample distribution across failure modes. As a result, for the five critical failure modes $F_1$, $F_2$, $F_3$, $F_4$, and $F_7$, a total of 242, 200, 199, 237, and 180 samples are generated, respectively. Also, note that 250 samples corresponding to the non-failure (safe) case are selected from MCS results. These results confirm that the proposed framework provides an effective distribution of samples across critical failure modes, including rare scenarios, addressing the limitations of the imbalanced dataset shown in Figure~\ref{Fig_Failure_mode_MCS}.

\subsection{DNN prediction: balanced vs. imbalanced datasets} \label{DNN_results}

To evaluate the effectiveness of the balanced dataset, two distinct DNN models are trained using both balanced and imbalanced datasets, respectively, based on the proposed architecture shown in Figure~\ref{Fig_DNN}. The input and output dimensions of the DNN models are $11 \times 1$ and $3 \times 1$, respectively.  

To mitigate the issue of overfitting, we employ different numbers of units in the hidden layers while maintaining the same number of hidden layers (i.e., three hidden layers) for the two DNN models. Specifically, the DNN model trained with the imbalanced dataset consists of 16, 8, and 4 units in its hidden layers, whereas the model trained with the balanced dataset consists of 32, 16, and 8 units. Each dataset is first split into training and test subsets, with 80\% allocated for training and 20\% for testing. The DNN models are trained for 2,000 epochs using the training dataset by minimizing the binary cross-entropy loss function with the Adam optimizer.

Tables~\ref{Tab_imbalanced} and ~\ref{Tab_balanced} present the prediction accuracy of the trained DNN models. Table~\ref{Tab_imbalanced} reports the accuracy of the model trained on the imbalanced dataset, while Table~\ref{Tab_balanced} shows the accuracy of the model trained on the balanced dataset. To further evaluate model performance, each DNN model is also tested on the opposite dataset. In particular, Table~\ref{Tab_imbalanced} includes results for the imbalanced model tested on the balanced dataset, whereas Table~\ref{Tab_balanced} shows the performance of the balanced model tested on the imbalanced dataset.

Two key observations are found from these results. First, there is no strong evidence of overfitting in either model. Second, the model trained on the balanced dataset demonstrates higher accuracy compared to the one trained on the imbalanced dataset. Notably, the imbalanced model performs poorly when tested on the balanced dataset, whereas the balanced model maintains comparable accuracy, as shown in Table~\ref{Tab_balanced}, when tested on the imbalanced dataset. This discrepancy arises because the majority of samples in the imbalanced dataset correspond to the safe case [0 0 0], leading to a common misclassification where the model predicts [0 0 0] even when the true label indicates failure.

\begin{table}[H]
  \caption{\textbf{Prediction accuracy of DNN model trained using imbalanced dataset}.}
  \label{Tab_imbalanced}
  \centering
  \begin{tabular}{c c c}
    \toprule
    & \textbf{Imbalanced dataset} & \textbf{Balanced dataset} \\
    \midrule
    Training set & 88.8\% & 28.7\% \\
    Test set  & 85.7\% & 24.4\% \\
    \bottomrule
  \end{tabular}
\end{table}
\begin{table}[H]
  \caption{\textbf{Prediction accuracy of DNN model trained using balanced dataset}.}
  \label{Tab_balanced}
  \centering
  \begin{tabular}{c c c}
    \toprule
    & \textbf{Imbalanced dataset} & \textbf{Balanced dataset} \\
    \midrule
    Training set & 87.6\% & 92.4\% \\
    Test set  & 87.6\% & 89.7\% \\
    \bottomrule
  \end{tabular}
\end{table}
Since the objective of this examination is to demonstrate the effectiveness of the balanced dataset instead of proposing a highly optimized DNN architecture for precise failure mode prediction, a simple architecture is sufficient. However, if the goal is to achieve higher predictive accuracy, one could improve performance by utilizing the entire acceleration history and incorporating advanced deep learning models such as transformers \cite{vaswani2017attention}.

\section{Numerical investigations} \label{Examples}

The accuracy and effectiveness of the proposed balanced dataset construction framework are demonstrated through numerical investigations of two finite element structural models: a nine-story steel building and a three-story moment-resisting frame (MRF) structure. The nine-story steel building is introduced to illustrate the framework's applicability to more complex structural systems compared to the three-story shear-type structure discussed in the previous section. Additionally, the framework is applied to the three-story MRF structure subjected to a synthetic ground motion set, demonstrating not only its applicability to high-fidelity numerical models but also its robustness across different ground motion datasets.

\subsection{Nine-story steel building structure} \label{Ex1:9SAC}

\subsubsection{Structural model and random variables}

\begin{figure}[H]
  \centering
  \includegraphics[scale=0.41] {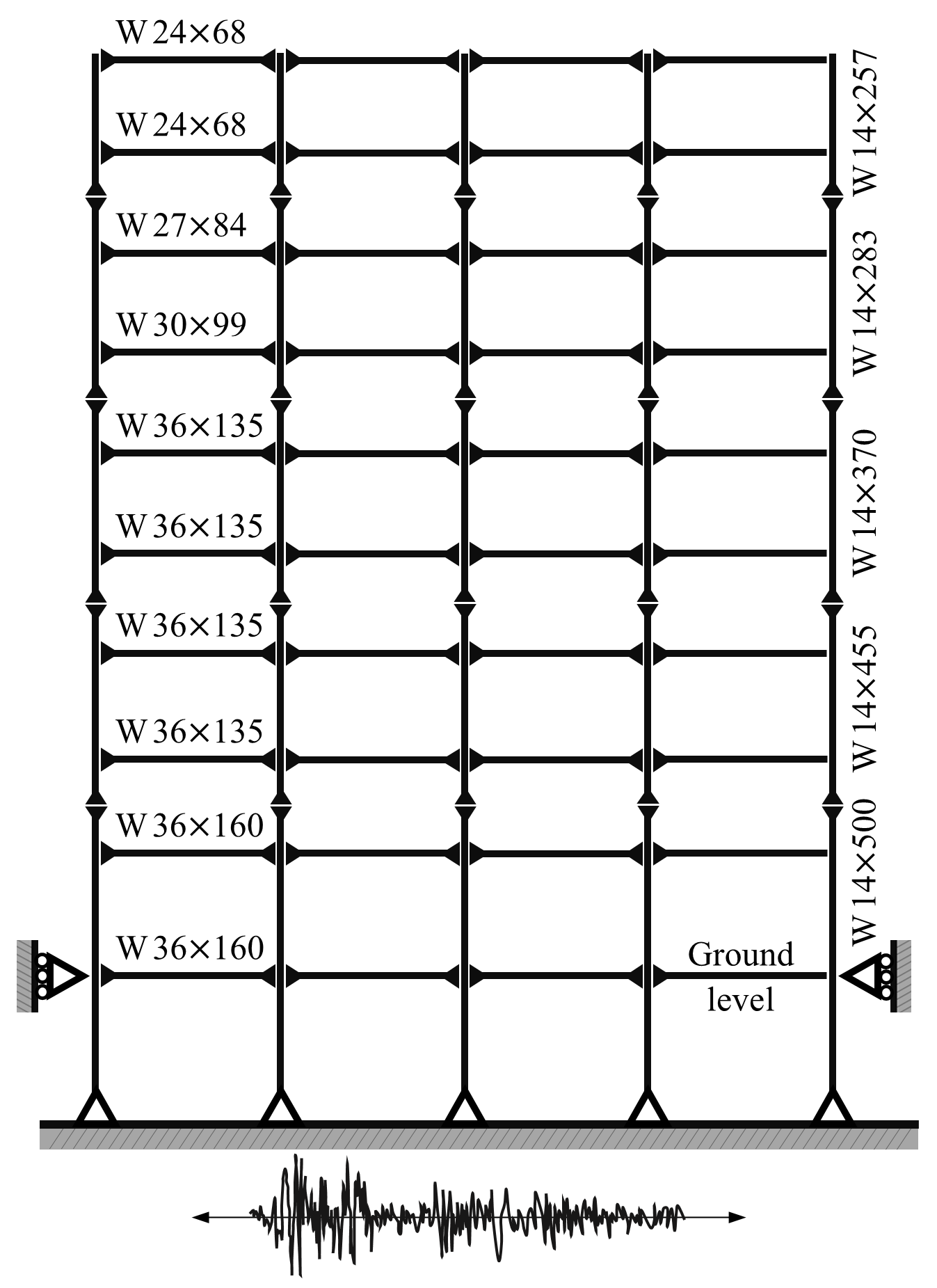}
  \caption{\textbf{Nine-story steel building model}.}
  \label{Fig_SAC9}
\end{figure}

A nine-story steel building structure, adopted from the SAC Phase II steel project \cite{ohtori2004benchmark} is modeled in OpenSees \cite{mckenna2011opensees}. Figure~\ref{Fig_SAC9} illustrates the configuration of the structure, which has a width of 45.73 m and a total height of 37.19 m. Column splices are located at the first, third, fifth, and seventh levels to resist uplift forces during seismic excitations, and the basement is laterally restrained to prevent horizontal displacement. Concrete foundation walls and surrounding soil provide additional lateral confinement at the base of the structure. To capture the nonlinear force-displacement behavior of beams and columns, a bilinear material model is employed. Furthermore, a fiber model is utilized to define the cross-sections of beams and columns.

Six structural parameters summarized in Table~\ref{Tab_9SACrvs} are treated as random variables. Considering the rigid diaphragm effect, each story is treated as an individual component when defining failure modes. Treating each story as a component is widely adopted in structural resilience analysis research \cite{kim2024accelerated,kim2024efficient}. Component failure is defined as the IDR exceeding the threshold of 1.5\%. Given that the building structure consists of nine components (i.e., 9 stories), the total number of potential failure modes is $N_F=511$ ($2^9-1$).

\begin{table}[H]
  \caption{\textbf{Probabilistic distributions of uncertain parameters for the nine-story steel building}.}
  \label{Tab_9SACrvs}
  \centering
  \begin{tabular}{c c c c}
    \toprule
    Variable & Mean & CoV & Distribution \\
    \midrule
    Damping ratio (\%) & 3 & 0.20 & Lognormal \\
    Modulus of elasticity (MPa) & 200,000 & 0.10 & Lognormal \\
    Yield strength for beam (MPa) & 248 & 0.15 & Lognormal \\
    Yield strength for column (MPa) & 345 & 0.15 & Lognormal \\
    Strain hardening ratio for beam & 0.01 & 0.25 & Lognormal \\
    Strain hardening ratio for column & 0.01 & 0.25 & Lognormal \\
    \bottomrule
  \end{tabular}
\end{table}

\subsubsection{Failure mode identification and balanced dataset construction}

\begin{figure}[H]
  \centering
  \includegraphics[scale=0.55] {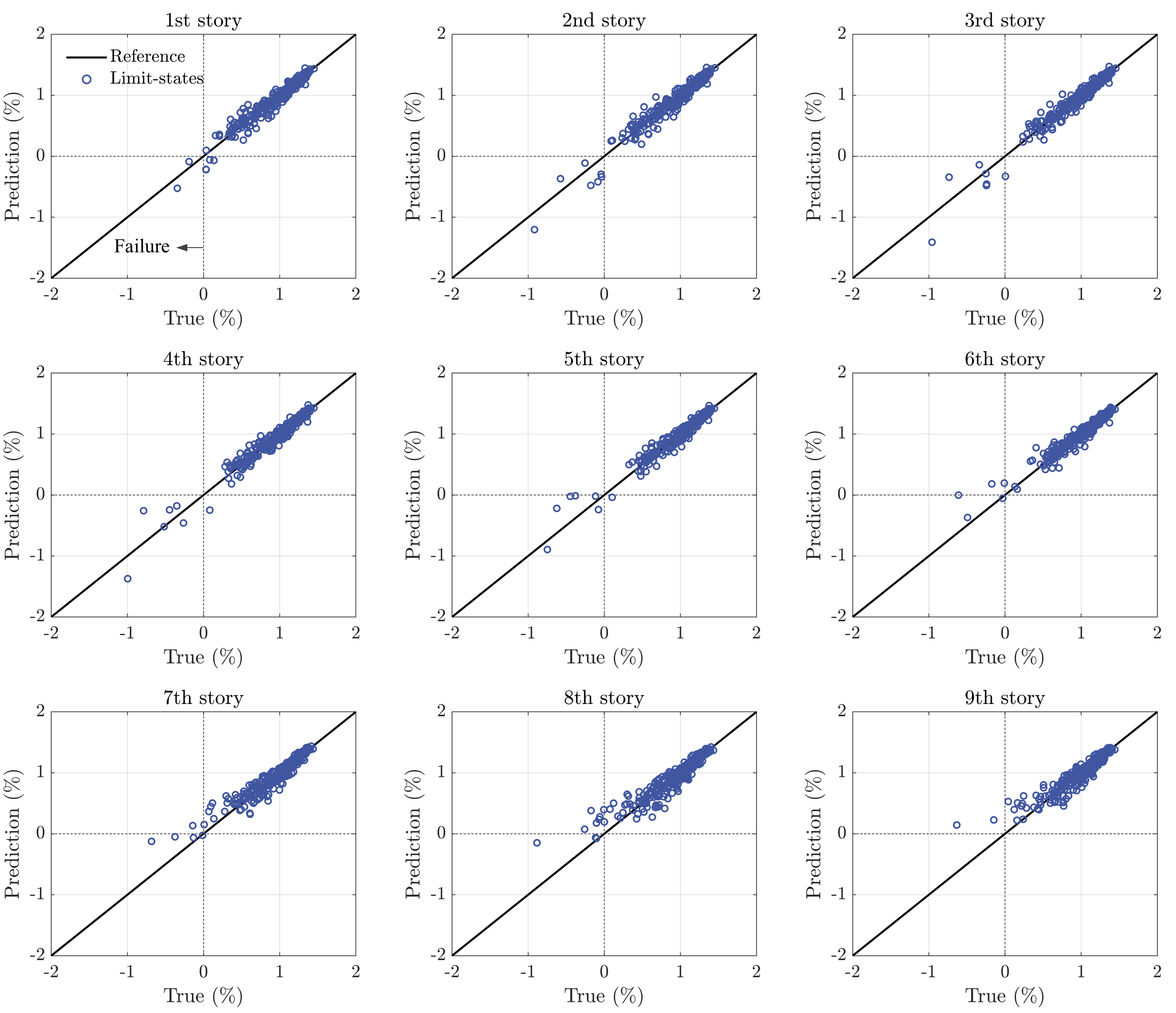}
  \caption{\textbf{Scatter plots of true limit state values against GP predictions for the nine-story steel building}. The solid line represents perfect agreement between predictions and observations, while the dashed line indicates the failure threshold.}
  \label{Fig_GP_scatter_9SAC}
\end{figure}

The 523 spectrum-compatible ground motions introduced in Section~\ref{GM_and_MDOF} are employed along with the eight critical GMFs from Section~\ref{GMF_selection}, to characterize aleatoric uncertainties. Following the procedure described in Section~\ref{FailMode_Algo}, GP surrogate models are initially trained using 150 initial samples. Subsequently, 65 additional samples are adaptively selected to refine the GP models. Figure~\ref{Fig_GP_scatter_9SAC} illustrates the scatter plots of true limit state values compared to their GP predictions, which demonstrates high accuracy. The refined models achieve $R^2$ values of 0.9355, 0.9370, 0.9445, 0.9489, 0.9439, 0.9282, 0.9042, 0.8760, and 0.8643 for the first through ninth story limit states, respectively.

By setting the algorithm parameters as $n_{b}=10^7$, $R=5.0$ and $n_m=3$, the framework identifies 11 critical failure modes, as summarized in Table~\ref{Tab_FM_9SAC}. For simplicity, the mode indices in the table represent combinations of failed components instead of a numerical sequence. For example, $F_{7,8}$ denotes a failure mode where only the seventh and eighth stories fail, while other stories remain intact. Due to the strong correlation of story drifts in building structures, multi-story failures are observed more frequently than single- or bi-component failures.  A balanced dataset is constructed by uniformly generating 250 samples for each failure mode, resulting in a total of 2,750 samples. These samples are transformed into the ground motion time history domain through the scaling factor optimization process described in Section~\ref{Construction}.

\begin{figure}[H]
  \centering
  \includegraphics[scale=0.53] {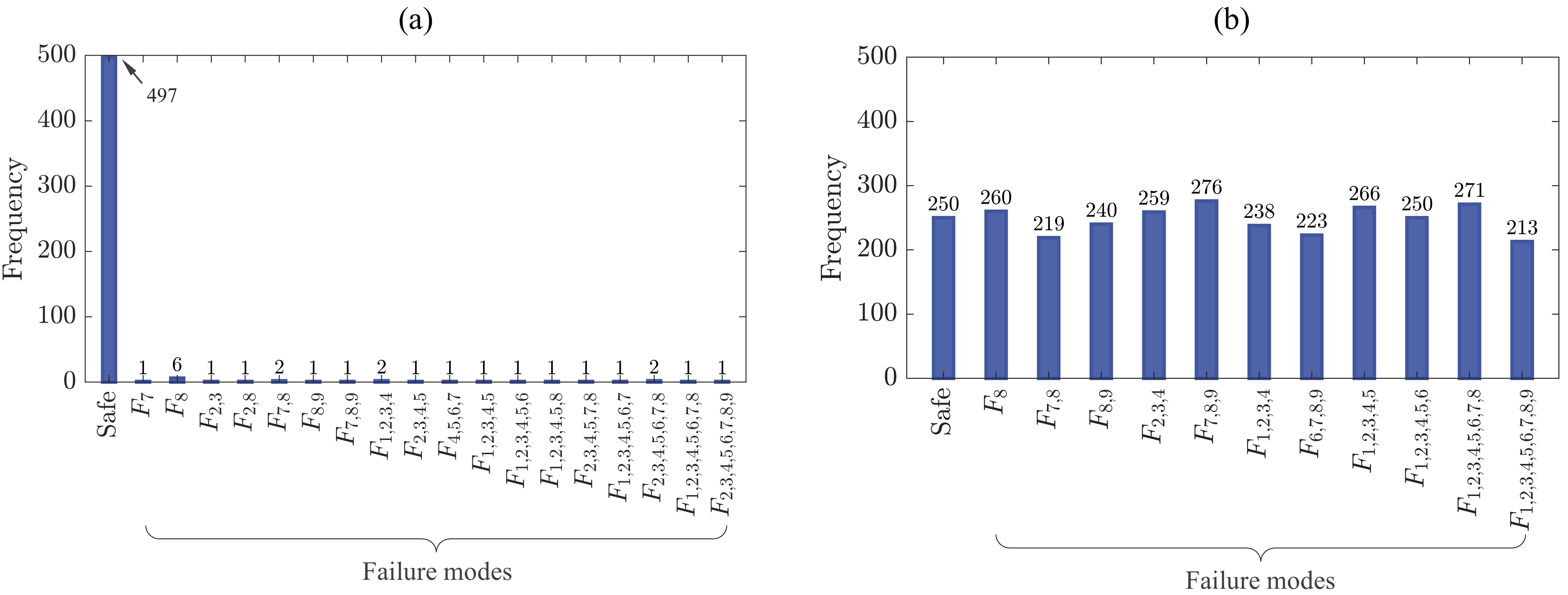}
  \caption{\textbf{Comparison of failure mode distributions for the nine-story steel building: (a) imbalanced dataset from MCS and (b) balanced dataset generated by the proposed framework}.}
  \label{Fig_Failure_mode_9SAC}
\end{figure}

Figure~\ref{Fig_Failure_mode_9SAC} compares the failure mode distributions from the imbalanced dataset (obtained via MCS) and the balanced dataset (generated by the proposed framework). For clarity, Figure~\ref{Fig_Failure_mode_9SAC}(a) includes only failure modes that occurred at least once in the MCS results. The MCS dataset is heavily skewed toward non-failure cases, with only a small fraction of samples representing various failure modes. In contrast, the proposed framework identifies 11 critical failure modes and constructs a balanced dataset that provides equitable representation across these modes. While the MCS dataset includes 18 failure modes, the additional modes are rare events outside the $n$-ball domain with occurrence probabilities below $\Phi(-5) = 2.8\times10^{-7}$ and are excluded as non-critical in the balanced dataset. Additionally, 250 samples corresponding to the non-failure (safe) case are included from the MCS results in the balanced dataset. The findings demonstrate that the proposed framework effectively addresses the data imbalance issue commonly encountered in traditional MCS-based approaches, particularly for high-fidelity structural models with a large number of potential component failures.

\begin{table}[H]
  \caption{\textbf{Failure modes identified for the nine-story steel building}.}
  \label{Tab_FM_9SAC}
  \centering
  \begin{tabular}{c c c}
    \toprule
    Case & Mode index ($i$) & Failure mode ($F_i$) \\
    \midrule
    1 & 8 & $\left\{\overline{C}_1 \overline{C}_2 \overline{C}_3 \overline{C}_4 \overline{C}_5 \overline{C}_6 \overline{C}_7 C_8 \overline{C}_{9} \right\}$ \\
    2 & 7,8 & $\left\{\overline{C}_1 \overline{C}_2 \overline{C}_3 \overline{C}_4 \overline{C}_5 \overline{C}_6 C_7 C_8 \overline{C}_{9} \right\}$ \\
    3 & 8,9 & $\left\{\overline{C}_1 \overline{C}_2 \overline{C}_3 \overline{C}_4 \overline{C}_5 \overline{C}_6 \overline{C}_7 C_8 C_{9} \right\}$ \\
    4 & 2,3,4 & $\left\{\overline{C}_1 C_2 C_3 C_4 \overline{C}_5 \overline{C}_6 \overline{C}_7 \overline{C}_8 \overline{C}_{9} \right\}$ \\
    5 & 7,8,9 & $\left\{\overline{C}_1 \overline{C}_2 \overline{C}_3 \overline{C}_4 \overline{C}_5 \overline{C}_6 C_7 C_8 C_{9} \right\}$ \\
    6 & 1,2,3,4 & $\left\{C_1 C_2 C_3 C_4 \overline{C}_5 \overline{C}_6 \overline{C}_7 \overline{C}_8 \overline{C}_{9} \right\}$ \\
    7 & 6,7,8,9 & $\left\{\overline{C}_1 \overline{C}_2 \overline{C}_3 \overline{C}_4 \overline{C}_5 C_6 C_7 C_8 C_{9} \right\}$ \\
    8 & 1,2,3,4,5 & $\left\{C_1 C_2 C_3 C_4 C_5 \overline{C}_6 \overline{C}_7 \overline{C}_8 \overline{C}_{9} \right\}$ \\
    9 & 1,2,3,4,5,6 & $\left\{C_1 C_2 C_3 C_4 C_5 C_6 \overline{C}_7 \overline{C}_8 \overline{C}_{9} \right\}$ \\
    10 & 1,2,3,4,5,6,7,8 & $\left\{C_1 C_2 C_3 C_4 C_5 C_6 C_7 C_8 \overline{C}_{9} \right\}$ \\
    11 & 1,2,3,4,5,6,7,8,9 & $\left\{C_1 C_2 C_3 C_4 C_5 C_6 C_7 C_8 C_9 \right\}$ \\
    \bottomrule
  \end{tabular}
\end{table}

\subsubsection{DNN model performance: balanced vs. imbalanced datasets}
Due to the presence of nine components in the structure, the input and output dimensions are adjusted to $17 \times 1$ and $9 \times 1$, respectively. Furthermore, the number of units in the hidden layers of the DNN model trained on the balanced dataset is configured as 64, 32, and 16 for the first, second, and third hidden layers, respectively. The DNN models are trained on both imbalanced and balanced datasets using the same training environments described in Section~\ref{DNN_results}. Tables~\ref{Tab_9SAC_imbalanced} and~\ref{Tab_9SAC_balanced} present the accuracy of the DNN model trained on the imbalanced and balanced datasets, respectively. The prediction accuracy of the DNN models when tested on the opposite dataset is also reported.

The prediction accuracy of the DNN model trained on the imbalanced dataset decreases to approximately $1/12$ when tested on the balanced dataset. This result is inversely proportional to the number of safe cases and failure modes represented in the balanced dataset. The primary reason for this is that the DNN model trained on the imbalanced dataset tends to predict the safe mode regardless of the input data. Furthermore, it is observed that the model classifies every failure mode as a safe case within the imbalanced dataset. This occurs because the safe case constitutes approximately 95\% of the dataset, leading the model to favor the majority class and potentially overlook minority class instances. 

In contrast, the DNN model trained on the balanced dataset demonstrates better prediction accuracy across both datasets over the other DNN model. Furthermore, its accuracy in predicting failure modes in the balanced dataset reaches approximately 46\%, confirming that a model trained on a balanced dataset is more effective in identifying failure modes with extremely low probabilities of occurrence. The lower prediction accuracy of 46\%, compared to the value reported in the third row of Table~\ref{Tab_9SAC_balanced}, can be attributed to the presence of several failure modes in the imbalanced dataset that are not included in the balanced dataset.

However, given the prediction accuracy of the DNN model trained on the balanced dataset, the input data used to develop the DNN model may not be insufficient for accurately predicting failure modes, particularly in structures with a substantial number of components. The primary reason for this is that the number of failure modes increases exponentially as the number of components grows, requiring more information to properly predict the state of each component. Therefore, further research is required to systematically determine the optimal input features for predicting failure modes, especially in such complex cases. Note that the objective of this study is to highlight the importance of using a balanced dataset over an imbalanced one, rather than to propose the optimal input dataset and architecture for failure mode prediction.

\begin{table}[H]
  \caption{\textbf{Prediction accuracy of DNN model trained using imbalanced dataset for the nine-story steel building}.}
  \label{Tab_9SAC_imbalanced}
  \centering
  \begin{tabular}{c c c}
    \toprule
    & \textbf{Imbalanced dataset} & \textbf{Balanced dataset} \\
    \midrule
    Training set & 95.2\% & 9.3\% \\
    Test set  & 94.3\% & 8.2\% \\
    \bottomrule
  \end{tabular}
\end{table}
\begin{table}[H]
  \caption{\textbf{Prediction accuracy of DNN model trained using balanced dataset for the nine-story steel building}.}
  \label{Tab_9SAC_balanced}
  \centering
  \begin{tabular}{c c c}
    \toprule
    & \textbf{Imbalanced dataset} & \textbf{Balanced dataset} \\
    \midrule
    Training set & 94.5\% & 76.9\% \\
    Test set  & 92.4\%  & 71.3\% \\
    \bottomrule
  \end{tabular}
\end{table}

\subsection{Three-story steel MRF structure subjected to synthetic ground motions} \label{Ex1:3SAC_synthetic}

\begin{figure}[H]
  \centering
  \includegraphics[scale=0.57] {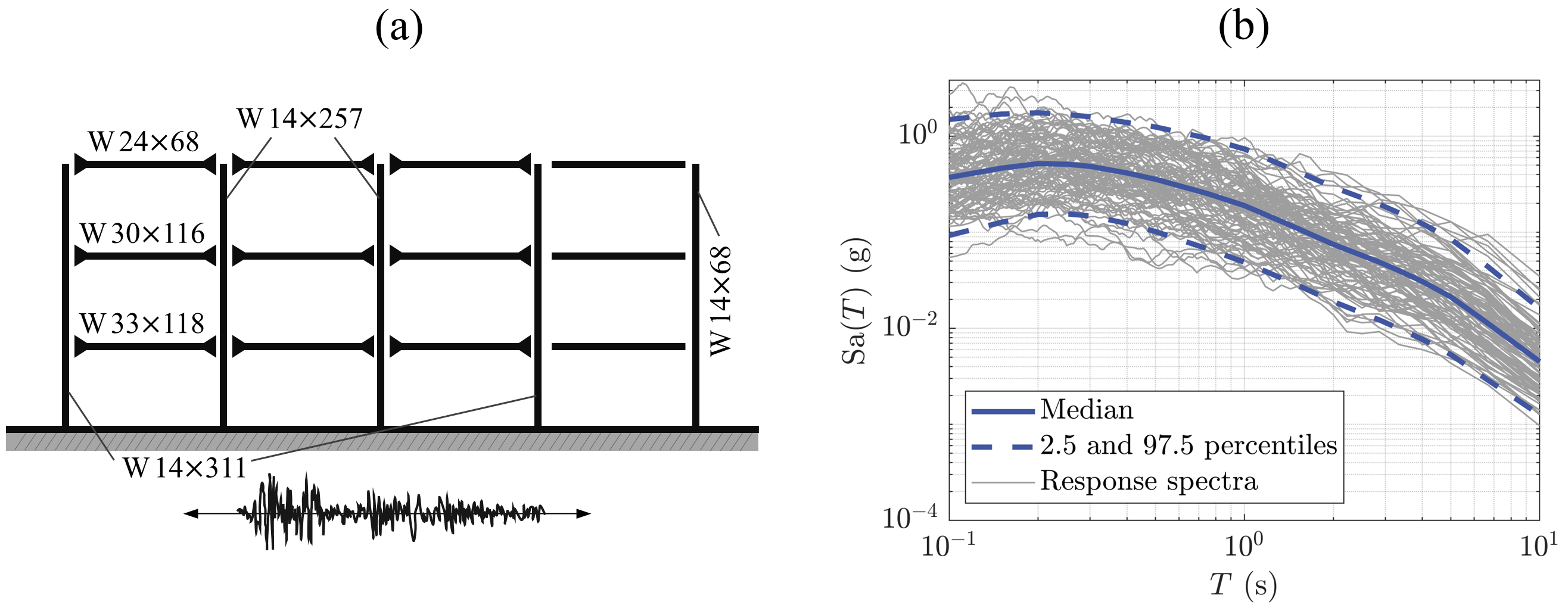}
  \caption{\textbf{(a) Three-story steel MRF model and (b) response spectra of generated synthetic ground motions with the target spectrum}. Figure (b) shows 100 response spectra with the median and 2.5\%-97.5\% quantiles of the target spectrum superimposed.}
  \label{Fig_SAC3_spectra}
\end{figure}

The second numerical investigation evaluates the framework's performance using synthetic ground motions, demonstrating its robustness to different sets of ground motions and structural systems. To this end, a three-story steel MRF structure, adopted from the SAC joint venture project \cite{ohtori2004benchmark}, is modeled in OpenSees. Figure~\ref{Fig_SAC3_spectra}(a) illustrates the configuration of the three-story building structure. Similar to the previous numerical investigation, the "uniaxialMaterial" command is utilized to capture the material nonlinearity, while the fiber section approach is employed for section modeling. Further details can be found in Ohtori et al.\cite{ohtori2004benchmark}.

Synthetic motions are generated using the stochastic ground motion model developed by Yanni et al.\cite{yanni2024probabilistic}, which simulates a filtered stochastic process modulated by two envelope functions and incorporates baseline correction for zero residual velocity and displacement. Similar to the recorded motion case, these motions are calibrated to a target spectrum following Section~\ref{GM_and_MDOF} with seismic hazard parameters corresponding to an earthquake magnitude of 6.5, rupture distance of 10 km, average shear wave velocity of 450 m/s, a normal faulting mechanism, and the California region. A total of 1,000 synthetic ground motions are generated, and their response spectra are presented in Figure~\ref{Fig_SAC3_spectra}(b).

Similar to Section~\ref{Ex1:9SAC}, each story is treated as an individual component in defining the failure mode. Structural component failure is characterized by IDR exceeding the threshold of 1.7\%, resulting in $N_F=7$ possible failure modes. Epistemic uncertainties in structural parameters are summarized in Table~\ref{Tab_3SACrvs}, while the eight critical GMFs are employed to account for aleatoric uncertainties.
\begin{table}[H]
  \caption{\textbf{Probabilistic distributions of uncertain parameters for the three-story MRF structure}.}
  \label{Tab_3SACrvs}
  \centering
  \begin{tabular}{c c c c}
    \toprule
    Variable & Mean & CoV & Distribution \\
    \midrule
    Damping ratio (\%) & 3 & 0.20 & Lognormal \\
    Modulus of elasticity (MPa) & 200000 & 0.10 & Lognormal \\
    Yield strength for beam (MPa) & 248 & 0.15 & Lognormal \\
    Yield strength for column (MPa) & 345 & 0.15 & Lognormal \\
    Strain hardening ratio for beam & 0.01 & 0.25 & Lognormal \\
    Strain hardening ratio for column & 0.01 & 0.25 & Lognormal \\
    \bottomrule
  \end{tabular}
\end{table}

\begin{figure}[H]
  \centering
  \includegraphics[scale=0.48] {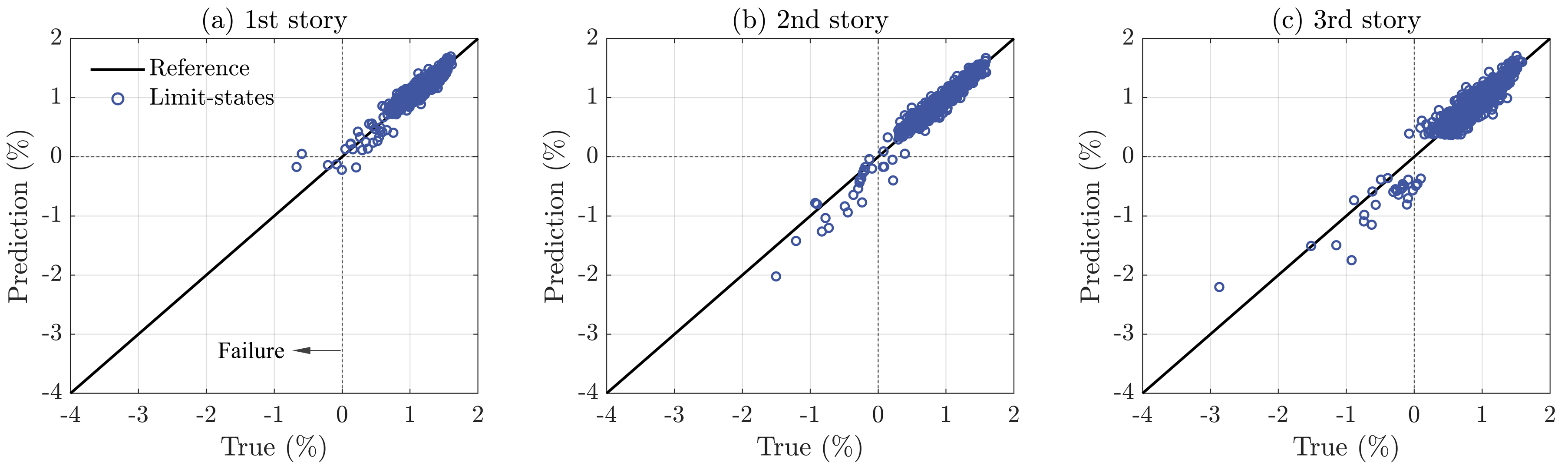}
  \caption{\textbf{Scatter plots of true limit state values against GP predictions for the three-story MRF structure: (a) 1st story, (b) 2nd story, and (c) 3rd story}. The solid line represents perfect agreement between predictions and observations, while the dashed line indicates the failure threshold.}
  \label{Fig_GP_scatter_3SAC}
\end{figure}

The proposed framework is initialized with 200 training samples, followed by the adaptive selection of 109 additional points to identify the probability density of each failure mode. Figure~\ref{Fig_GP_scatter_3SAC} presents scatter plots of GP predictions against true limit state values, demonstrating high predictive accuracy with $R^2$ values of 0.8941, 0.9431, and 0.8687 for the first, second, and third story limit states, respectively. Using parameters of $n_{b}=10^7$, $R=6.0$, and $n_m=3$, the framework identifies five critical failure modes, as summarized in Table~\ref{Tab_FM_3SAC}.

\begin{figure}[H]
  \centering
  \includegraphics[scale=0.51] {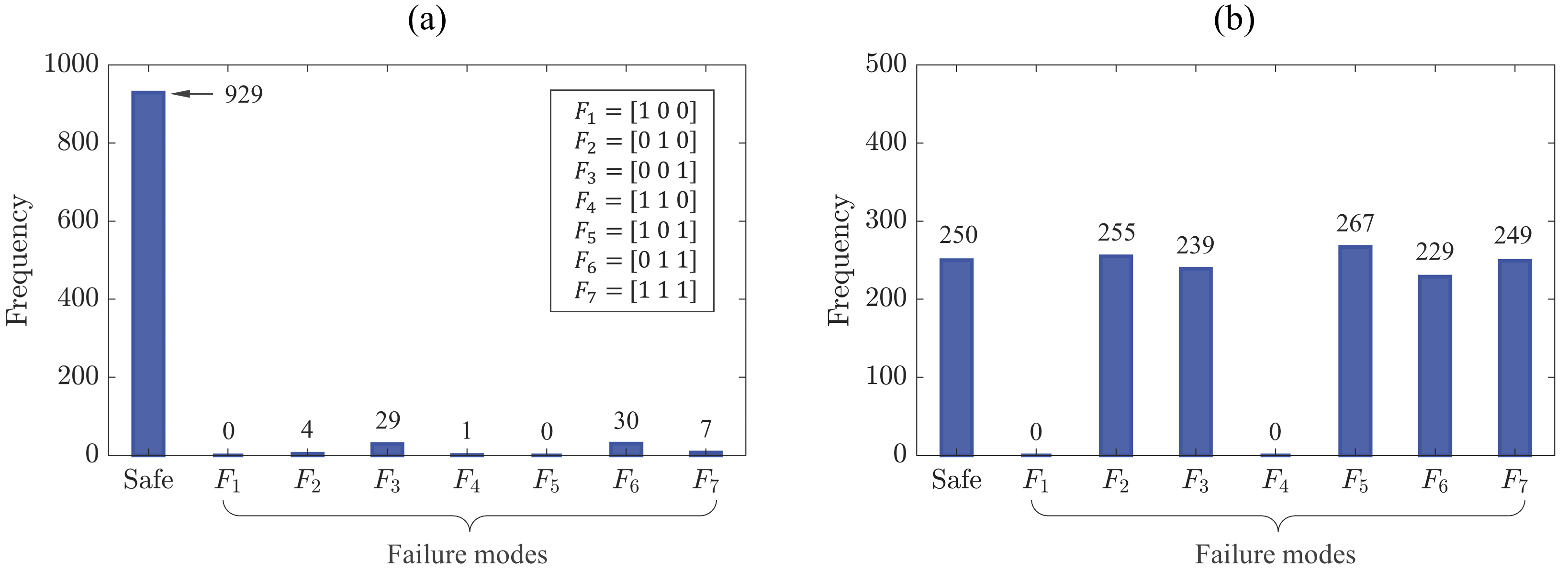}
  \caption{\textbf{Comparison of failure mode distributions for the three-story MRF structure: (a) imbalanced dataset from MCS and (b) balanced dataset generated by the proposed framework}.}
  \label{Fig_Failure_mode_3SAC}
\end{figure}

To construct balanced datasets, 250 samples are uniformly generated for each failure mode (a total of 1,250 samples) and transformed into the ground motion time history domain using the scaling factor optimization process. Figure~\ref{Fig_Failure_mode_3SAC} compares the failure mode distributions from the imbalanced dataset obtained via MCS with the balanced dataset generated by the proposed framework. In the balanced dataset, 250 non-failure (safe) samples are also included, extracted from the MCS results. These findings confirm the framework's capability to ensure improved representation across failure modes, addressing challenges posed by imbalanced datasets in traditional seismic analyses.
\begin{table}[H]
  \caption{\textbf{Failure modes identified for the three-story MRF structure}.}
  \label{Tab_FM_3SAC}
  \centering
  \begin{tabular}{c c c}
    \toprule
    Case & Mode index ($i$) & Failure mode ($F_i$) \\
    \midrule
    1 & 2 & $\left\{\overline{C}_1 C_2 \overline{C}_{3} \right\}$ \\
    2 & 3 & $\left\{\overline{C}_1 \overline{C}_2 C_{3} \right\}$ \\
    3 & 5 & $\left\{C_1 \overline{C}_2 C_{3} \right\}$ \\
    4 & 6 & $\left\{\overline{C}_1 C_2 C_{3} \right\}$ \\
    5 & 7 & $\left\{C_1 C_2 C_{3} \right\}$ \\
    \bottomrule
  \end{tabular}
\end{table}

Using the same architecture and training environments described in Section~\ref{DNN_results}, two DNN models are trained on the imbalanced and balanced datasets, respectively. As in previous investigations, Table~\ref{Tab_3SAC_imbalanced} presents the accuracy of the model trained on the imbalanced dataset, while Table~\ref{Tab_3SAC_balanced} reports the accuracy of the model trained on the balanced dataset. Additionally, each DNN model is tested on the opposite dataset. Similar results are observed, demonstrating that the proposed framework maintains consistent performance across different types of ground motions, with comparable effectiveness for both synthetic and recorded ground motion sets.

\begin{table}[H]
  \caption{\textbf{Prediction accuracy of DNN model trained using imbalanced dataset for the three-story MRF structure}.}
  \label{Tab_3SAC_imbalanced}
  \centering
  \begin{tabular}{c c c}
    \toprule
    & \textbf{Imbalanced dataset} & \textbf{Balanced dataset} \\
    \midrule
    Training set & 95.6\% & 25.7\% \\
    Test set  & 96.4\% & 23.1\% \\
    \bottomrule
  \end{tabular}
\end{table}
\begin{table}[H]
  \caption{\textbf{Prediction accuracy of DNN model trained using balanced dataset for the three-story MRF structure}.}
  \label{Tab_3SAC_balanced}
  \centering
  \begin{tabular}{c c c}
    \toprule
    & \textbf{Imbalanced dataset} & \textbf{Balanced dataset} \\
    \midrule
    Training set & 98.1\% & 95.2\%  \\
    Test set  & 97.5\% & 93.8\%  \\
    \bottomrule
  \end{tabular}
\end{table}

\section{Conclusions} \label{Conclusion}

This study proposed a framework for constructing a balanced dataset to facilitate training machine learning (or deep learning) models for predicting structural failure modes under seismic excitations. The framework achieved this by introducing three key methodologies. First, a procedure was developed to identify critical ground motion features (GMFs) that effectively capture the variability in structural responses. Second, an adaptive algorithm based on Gaussian processes was designed to efficiently estimate the probability densities of failure modes while accounting for both epistemic and aleatoric uncertainties. Third, a scaling factor optimization process was implemented to transform generated samples from the critical GMF space to the ground motion time history domain without compromising the physical characteristics of seismic excitations. The framework’s effectiveness was demonstrated by training deep neural network (DNN) models on both balanced and imbalanced datasets. Two numerical investigations including nine-story and three-story building structures validated the framework’s applicability to complex structural systems and its robustness across different sets of ground motions. Results consistently indicate higher prediction accuracy for models trained on the balanced dataset, highlighting the merits of the proposed framework in improving machine learning performance for structural failure mode prediction. Further research includes two key areas. First, while this study employed a simple DNN architecture for classifying the failure modes, future research can explore the integration of advanced deep learning models, such as transformers \cite{zhang2024transformer} or graph neural networks \cite{kim2025near}, to further improve the prediction accuracy. Second, integrating the framework with reliability-based design optimization \cite{kim2021quantile} and resilience assessment \cite{yi2023system} can broaden its utility in risk-informed decision-making and seismic resilience analysis.

\section*{Acknowledgments}
This research was supported by the National Research Foundation of Korea grant funded by the Korea government (MSIT) (RS-2023-00242859). This research was also supported by the Basic Science Research Program through the National Research Foundation of Korea funded by the Ministry of Education (RS-2024-00407901). These supports are gratefully acknowledged.

\appendix
\renewcommand{\theequation}{A.\arabic{equation}}
\renewcommand{\thefigure}{A.\arabic{figure}}
\renewcommand{\thetable}{A.\arabic{table}}
\setcounter{figure}{0} 
\setcounter{table}{0} 

\section{Fundamentals of GP modeling} \label{App:GP}

A Gaussian process (GP) provides a probabilistic framework for modeling responses $f(\mathbf{x})$ as realizations of a stochastic process \cite{williams2006gaussian}. The GP model is defined as:
\begin{equation} \label{Eq:GP_model}
f(\mathbf{x}) \sim GP\left(\mu_{f}(\mathbf{x}),k_f\left({\mathbf{x},\mathbf{x}';\mathbf{\theta}}\right)\right)   \,,
\end{equation}
where $\mu_{f}(\mathbf{x})$ is the mean function, $k_f(\mathbf{x},\mathbf{x}')$ is the covariance kernel function, and $\mathbf{\theta}$ represents hyperparameters defining the kernel's structure.

In practice, observations often include Gaussian noise, leading to $y=f(\mathbf{x})+\varepsilon$, where $\varepsilon \sim N(0,\sigma_n^2)$. Given $n$ training samples $\mathcal{X}_{\mathcal{D}}=[\mathbf{x}_1,...,\mathbf{x}_n]^T$ and corresponding observations $\mathcal{Y}_{\mathcal{D}}=[y_1,...,y_n]^T$, the hyperparameters $\mathbf{\theta}$ can be estimated by maximizing the log marginal likelihood:
\begin{equation} \label{Eq:GP_like}
\ln{p(\mathcal{Y}_{\mathcal{D}} | \mathcal{X}_{\mathcal{D}},\mathbf{\theta})} = -\frac{1}{2}\mathcal{Y}_{\mathcal{D}}^T(\mathbf{K_f} + \sigma_n^2\mathbf{I})^{-1}\mathcal{Y}_{\mathcal{D}} - \frac{1}{2} \ln{|\mathbf{K_f} + \sigma_n^2\mathbf{I}|} - \frac{n}{2} \ln{2\pi}
\,,
\end{equation}
where $\mathbf{K_f}$ is the covariance matrix with elements $K_{f_{ij}}=k_f(\mathbf{x}_i,\mathbf{x}_j)$, and $\mathbf{I}$ is the identity matrix.

Once trained, the GP model predicts the response at a new input $\mathbf{x}_*$. The predictive mean and variance are:
\begin{equation} \label{Eq:GP_mean}
\mu_{\hat{Y}}\left(\mathbf{x}_*\right) = \mathbf{k}_{f_*}^T(\mathbf{K_f} + \sigma_n^2\mathbf{I})^{-1}\mathcal{Y}_{\mathcal{D}}     \,,
\end{equation}
\begin{equation} \label{Eq:GP_var}
\sigma^2_{\hat{Y}}\left(\mathbf{x}_*\right) = k_{f_{**}} - \mathbf{k}_{f_*}^T(\mathbf{K_f} + \sigma_n^2\mathbf{I})^{-1}\mathbf{k}_{f_*}     \,,
\end{equation}
where $\mathbf{k}_{f_*}$ is the covariance vector between $\mathbf{x}_*$ and the training points, and $k_{f_{**}}$ is the self-covariance of $\mathbf{x}_*$. The GP model not only provides predictions through $\mu_{\hat{Y}}\left(\mathbf{x}_*\right)$ but also quantifies the uncertainty via $\sigma^2_{\hat{Y}}\left(\mathbf{x}_*\right)$, making it suitable for applications requiring uncertainty-aware modeling and active learning.

\bibliography{Failure_mode}

\end{document}